\documentclass[11pt,a4paper]{article}
\usepackage[hyperref]{acl2021}
\usepackage{times}
\usepackage{latexsym}

\usepackage{microtype}

\usepackage{booktabs}
\usepackage{multirow}
\usepackage{amsmath}
\usepackage{amssymb}
\usepackage{xcolor}
\usepackage{pifont}

\usepackage{tikz}
\usepackage{subfigure}
\usepackage{pgfplots}
\usepackage{makecell,rotating,diagbox}

\pgfplotsset{compat=1.14}

\usetikzlibrary{plotmarks}
\usetikzlibrary{arrows}
\usetikzlibrary{decorations}
\usetikzlibrary{decorations.pathreplacing}
\usetikzlibrary{decorations.markings}
\usetikzlibrary{backgrounds}
\usetikzlibrary{fit}
\usetikzlibrary{positioning}
\usetikzlibrary{calc}
\usetikzlibrary{patterns}
\usetikzlibrary{shadows}
\usetikzlibrary[shapes.multipart]
\usetikzlibrary{pgfplots.groupplots}

\newdimen\base
\newdimen\baseh
\newdimen\basew
\baseh=\base
\basew=\base

\newdimen\legendmargin 
\newdimen\legendwidth 
\newdimen\legendsep 

\definecolor{ugreen}{rgb}{0,0.5,0}
\definecolor{ublue}{rgb}{0.152,0.250,0.545}
\definecolor{lyyblue}{RGB}{31,120,180}
\definecolor{lyygreen}{RGB}{51,160,44}
\definecolor{lyyred}{RGB}{227,26,28}
\definecolor{grayblue}{RGB}{214,220,229}
\definecolor{lightblue}{RGB}{157,195,230}
\definecolor{grayyellow}{RGB}{255,230,153}
\definecolor{lightgray}{RGB}{242,242,242}

\newcommand{\tab}[1]{Table \ref{#1}}%
\newcommand{\fig}[1]{Fig. \ref{#1}}%
\newcommand{\eqn}[1]{Eq. \ref{#1}}%

\newcommand{\tinystu}{\textsc{Tiny}}
\newcommand{\smallstu}{\textsc{Small}}

\DeclareMathOperator*{\argmin}{arg\,min}

\aclfinalcopy 

\title{Weight Distillation: Transferring the Knowledge\\ in Neural Network Parameters}

\author{
  Ye Lin\textsuperscript{1}\thanks{\ \ Authors contributed equally.}\ ,
  Yanyang Li\textsuperscript{2}$^{*}$,
  Ziyang Wang\textsuperscript{1},
  Bei Li\textsuperscript{1},
  Quan Du\textsuperscript{1},
  Tong Xiao\textsuperscript{1,3},
  Jingbo Zhu\textsuperscript{1,3}\thanks{\ \ Corresponding author.} \\
  \textsuperscript{1}NLP Lab, School of Computer Science and Engineering, \\
    Northeastern University, Shenyang, China \\
  \textsuperscript{2}The Chinese University of Hong Kong, Hong Kong, China \\
  \textsuperscript{3}NiuTrans Research, Shenyang, China \\
  {\tt \{linye2015,blamedrlee,libeineu,duquanneu\}@outlook.com}\\
  {\tt \{wangziyang\}@stumail.neu.edu.cn}\\
  {\tt \{xiaotong,zhujingbo\}@mail.neu.edu.cn} \\
}

\date{}

\begin{document}
\maketitle
\begin{abstract}
  Knowledge distillation has proven to be effective in model acceleration and compression. It transfers knowledge from a large neural network to a small one by using the large neural network predictions as targets of the small neural network. But this way ignores the knowledge inside the large neural networks, e.g., parameters. Our preliminary study as well as the recent success in pre-training suggests that transferring parameters are more effective in distilling knowledge. In this paper, we propose \emph{Weight Distillation} to transfer the knowledge in parameters of a large neural network to a small neural network through a parameter generator. On the WMT16 En-Ro, NIST12 Zh-En, and WMT14 En-De machine translation tasks, our experiments show that weight distillation learns a small network that is 1.88$\sim$2.94$\times$ faster than the large network but with competitive BLEU performance. When fixing the size of the small networks, weight distillation outperforms knowledge distillation by 0.51$\sim$1.82 BLEU points. The code is publicly available at https://github.com/Lollipop321/weight-distillation.
\end{abstract}

\section{Introduction}

Knowledge Distillation (KD) is a popular model acceleration and compression approach \cite{DBLP:journals/corr/HintonVD15}. It assumes that a lightweight network (i.e., \textit{student} network, or student for short) can learn to generalize in the same way as a large network (i.e., \textit{teacher} network, or teacher for short). To this end, a simple method is to train the student network with predicted probabilities of the teacher network as its targets.

But KD has its limitation: the student network can only access the knowledge in the predictions of the teacher network. It does not consider the knowledge in the teacher network parameters. These parameters contain billions of entries for the teacher network to make predictions. Yet in KD the student only learns from those predictions with at most thousands of categories. This way results in an inferior student network, since it learns from the limited training signals. Our analysis in Section \ref{sub-sec:initialization} shows that KD performs better if we simply cut off parts of parameters from the teacher to initialize the student. This fact implies that the knowledge in parameters is complementary to KD but missed. It also agrees with the recent success in pre-training \cite{DBLP:conf/nips/YangDYCSL19,DBLP:journals/corr/abs-1907-11692,DBLP:conf/naacl/DevlinCLT19}, where parameters reusing plays the main role. Based on this observation, a superior student is expected if all parameters in the teacher network could be exploited. However, this imposes a great challenge as the student network is too small to fit in the whole teacher network.

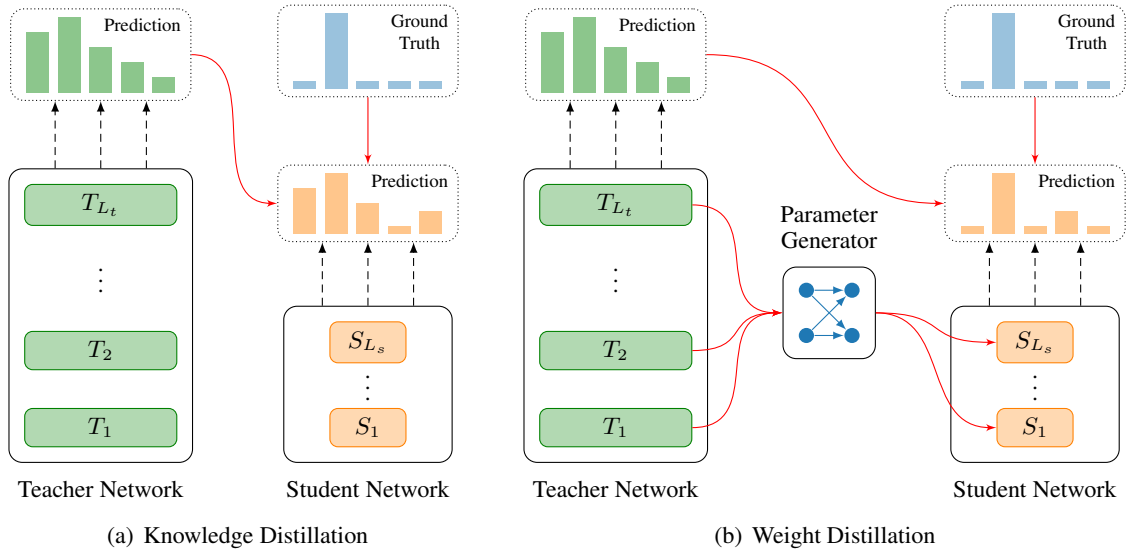
\begin{figure*}[t!]
  \centering
  \hspace*{\fill}
  \subfigure[Knowledge Distillation]
  {
      \begin{tikzpicture}
          \tikzstyle{weight} = [rectangle,minimum height=0.5cm,font=\small,rounded corners=3pt]
          \tikzstyle{teacher weight} = [weight,,draw=ugreen,fill=ugreen!30,minimum width=2cm]
          \tikzstyle{student weight} = [weight,draw=orange,fill=orange!30,minimum width=1cm]

          \tikzstyle{prob} = [rectangle,minimum width=0.3cm,inner sep=0pt]
          \tikzstyle{teacher prob} = [prob,fill=ugreen!45]
          \tikzstyle{student prob} = [prob,fill=orange!45]
          \tikzstyle{ground truth prob} = [prob,fill=lyyblue!45]

          \node[teacher weight] (tw1) at (0,0) {$T_{1}$};
          \node[teacher weight,anchor=south] (tw2) at ([yshift=0.5cm]tw1.north) {$T_{2}$};
          \node[font=\small,anchor=south,inner sep=0pt] (tds) at ([yshift=0.5cm]tw2.north) {$\vdots$};
          \node[teacher weight,anchor=south] (tw3) at ([yshift=0.35cm]tds.north) {$T_{L_t}$};

          \begin{pgfonlayer}{background}
            \node[draw,rounded corners=5pt,inner sep=0.2cm,fit=(tw1) (tw3),label={[font=\small,label distance=3pt]-90:Teacher Network}] (teacher) {};
          \end{pgfonlayer}

          \node[teacher prob,minimum height=0.6cm,anchor=south] (tp3) at ([yshift=1cm]teacher.north) {};
          \node[teacher prob,minimum height=1cm,anchor=south east] (tp2) at ([xshift=-0.1cm]tp3.south west) {};
          \node[teacher prob,minimum height=0.8cm,anchor=south east] (tp1) at ([xshift=-0.1cm]tp2.south west) {};
          \node[teacher prob,minimum height=0.4cm,anchor=south west] (tp4) at ([xshift=0.1cm]tp3.south east) {};
          \node[teacher prob,minimum height=0.2cm,anchor=south west] (tp5) at ([xshift=0.1cm]tp4.south east) {};

          \begin{pgfonlayer}{background}
            \coordinate (left) at ([xshift=-0.1cm]tp1.west);
            \coordinate (right) at ([xshift=0.1cm]tp5.east);
            \node[draw,densely dotted,rounded corners=5pt,inner sep=0.1cm,fit=(tp1) (tp2) (tp3) (tp4) (tp5) (left) (right)] (tp) {};
            \node[font=\scriptsize,align=center,anchor=north east] () at (tp.north east) {Prediction};
          \end{pgfonlayer}

          \node[student weight,anchor=west] (sw1) at ([xshift=2cm]tw1.east) {$S_{1}$};
          \node[font=\small,anchor=south,inner sep=0pt] (sds) at ([yshift=0.1cm]sw1.north) {$\vdots$};
          \node[student weight,anchor=south] (sw2) at ([yshift=-0.05cm]sds.north) {$S_{L_s}$};

          \begin{pgfonlayer}{background}
            \coordinate (left) at ([xshift=-0.4cm]sw2.west);
            \coordinate (right) at ([xshift=0.4cm]sw2.east);
            \node[draw,rounded corners=5pt,inner sep=0.2cm,fit=(sw1) (sw2) (left) (right),label={[font=\small,label distance=3pt]-90:Student Network}] (student) {};
          \end{pgfonlayer}

          \node[student prob,minimum height=0.4cm,anchor=south] (sp3) at ([yshift=0.95cm]student.north) {};
          \node[student prob,minimum height=0.8cm,anchor=south east] (sp2) at ([xshift=-0.1cm]sp3.south west) {};
          \node[student prob,minimum height=0.6cm,anchor=south east] (sp1) at ([xshift=-0.1cm]sp2.south west) {};
          \node[student prob,minimum height=0.1cm,anchor=south west] (sp4) at ([xshift=0.1cm]sp3.south east) {};
          \node[student prob,minimum height=0.3cm,anchor=south west] (sp5) at ([xshift=0.1cm]sp4.south east) {};

          \begin{pgfonlayer}{background}
            \coordinate (left) at ([xshift=-0.1cm]sp1.west);
            \coordinate (right) at ([xshift=0.1cm]sp5.east);
            \node[draw,densely dotted,rounded corners=5pt,inner sep=0.1cm,fit=(sp1) (sp2) (sp3) (sp4) (sp5) (left) (right)] (sp) {};
            \node[font=\scriptsize,align=center,anchor=north east] () at (sp.north east) {Prediction};
          \end{pgfonlayer}

          \node[ground truth prob,minimum height=0.1cm,anchor=south] (gp3) at ([yshift=1cm]sp.north) {};
          \node[ground truth prob,minimum height=1cm,anchor=south east] (gp2) at ([xshift=-0.1cm]gp3.south west) {};
          \node[ground truth prob,minimum height=0.1cm,anchor=south east] (gp1) at ([xshift=-0.1cm]gp2.south west) {};
          \node[ground truth prob,minimum height=0.1cm,anchor=south west] (gp4) at ([xshift=0.1cm]gp3.south east) {};
          \node[ground truth prob,minimum height=0.1cm,anchor=south west] (gp5) at ([xshift=0.1cm]gp4.south east) {};

          \begin{pgfonlayer}{background}
            \coordinate (left) at ([xshift=-0.1cm]gp1.west);
            \coordinate (right) at ([xshift=0.1cm]gp5.east);
            \node[draw,densely dotted,rounded corners=5pt,inner sep=0.1cm,fit=(gp1) (gp2) (gp3) (gp4) (gp5) (left) (right)] (gp) {};
            \node[font=\scriptsize,align=center,anchor=north east] () at (gp.north east) {Ground\\Truth};
          \end{pgfonlayer}

          \draw[-latex',red] (tp.east) .. controls +(east:1cm) and +(west:1cm) .. (sp.west);
          \draw[-latex',red] (gp.south) to (sp.north);

          \draw[-latex,densely dashed] (teacher.north) to (tp.south);
          \draw[-latex,densely dashed] ([xshift=-0.6cm]teacher.north) to ([xshift=-0.6cm]tp.south);
          \draw[-latex,densely dashed] ([xshift=0.6cm]teacher.north) to ([xshift=0.6cm]tp.south);

          \draw[-latex,densely dashed] (student.north) to (sp.south);
          \draw[-latex,densely dashed] ([xshift=-0.6cm]student.north) to ([xshift=-0.6cm]sp.south);
          \draw[-latex,densely dashed] ([xshift=0.6cm]student.north) to ([xshift=0.6cm]sp.south);
      \end{tikzpicture}
      \label{fig:kd}
  }
  \hfill
  \subfigure[Weight Distillation]
  {
      \begin{tikzpicture}
        \tikzstyle{weight} = [rectangle,minimum height=0.5cm,font=\small,rounded corners=3pt]
        \tikzstyle{teacher weight} = [weight,,draw=ugreen,fill=ugreen!30,minimum width=2cm]
        \tikzstyle{student weight} = [weight,draw=orange,fill=orange!30,minimum width=1cm]

        \tikzstyle{prob} = [rectangle,minimum width=0.3cm,inner sep=0pt]
        \tikzstyle{teacher prob} = [prob,fill=ugreen!45]
        \tikzstyle{student prob} = [prob,fill=orange!45]
        \tikzstyle{ground truth prob} = [prob,fill=lyyblue!45]

        \tikzstyle{pgnode} = [circle,fill=lyyblue,minimum size=0.2cm,inner sep=0pt]

        \node[teacher weight] (tw1) at (0,0) {$T_{1}$};
        \node[teacher weight,anchor=south] (tw2) at ([yshift=0.5cm]tw1.north) {$T_{2}$};
        \node[font=\small,anchor=south,inner sep=0pt] (tds) at ([yshift=0.5cm]tw2.north) {$\vdots$};
        \node[teacher weight,anchor=south] (tw3) at ([yshift=0.35cm]tds.north) {$T_{L_t}$};

        \begin{pgfonlayer}{background}
          \node[draw,rounded corners=5pt,inner sep=0.2cm,fit=(tw1) (tw3),label={[font=\small,label distance=3pt]-90:Teacher Network}] (teacher) {};
        \end{pgfonlayer}

        \node[teacher prob,minimum height=0.6cm,anchor=south] (tp3) at ([yshift=1cm]teacher.north) {};
        \node[teacher prob,minimum height=1cm,anchor=south east] (tp2) at ([xshift=-0.1cm]tp3.south west) {};
        \node[teacher prob,minimum height=0.8cm,anchor=south east] (tp1) at ([xshift=-0.1cm]tp2.south west) {};
        \node[teacher prob,minimum height=0.4cm,anchor=south west] (tp4) at ([xshift=0.1cm]tp3.south east) {};
        \node[teacher prob,minimum height=0.2cm,anchor=south west] (tp5) at ([xshift=0.1cm]tp4.south east) {};

        \begin{pgfonlayer}{background}
          \coordinate (left) at ([xshift=-0.1cm]tp1.west);
          \coordinate (right) at ([xshift=0.1cm]tp5.east);
          \node[draw,densely dotted,rounded corners=5pt,inner sep=0.1cm,fit=(tp1) (tp2) (tp3) (tp4) (tp5) (left) (right)] (tp) {};
          \node[font=\scriptsize,align=center,anchor=north east] () at (tp.north east) {Prediction};
        \end{pgfonlayer}

        \node[student weight,anchor=west] (sw1) at ([xshift=4cm]tw1.east) {$S_{1}$};
        \node[font=\small,anchor=south,inner sep=0pt] (sds) at ([yshift=0.1cm]sw1.north) {$\vdots$};
        \node[student weight,anchor=south] (sw2) at ([yshift=-0.05cm]sds.north) {$S_{L_s}$};

        \begin{pgfonlayer}{background}
          \coordinate (left) at ([xshift=-0.4cm]sw2.west);
          \coordinate (right) at ([xshift=0.4cm]sw2.east);
          \node[draw,rounded corners=5pt,inner sep=0.2cm,fit=(sw1) (sw2) (left) (right),label={[font=\small,label distance=3pt]-90:Student Network}] (student) {};
        \end{pgfonlayer}

        \node[student prob,minimum height=0.1cm,anchor=south] (sp3) at ([yshift=0.95cm]student.north) {};
        \node[student prob,minimum height=0.8cm,anchor=south east] (sp2) at ([xshift=-0.1cm]sp3.south west) {};
        \node[student prob,minimum height=0.1cm,anchor=south east] (sp1) at ([xshift=-0.1cm]sp2.south west) {};
        \node[student prob,minimum height=0.3cm,anchor=south west] (sp4) at ([xshift=0.1cm]sp3.south east) {};
        \node[student prob,minimum height=0.1cm,anchor=south west] (sp5) at ([xshift=0.1cm]sp4.south east) {};

        \begin{pgfonlayer}{background}
          \coordinate (left) at ([xshift=-0.1cm]sp1.west);
          \coordinate (right) at ([xshift=0.1cm]sp5.east);
          \node[draw,densely dotted,rounded corners=5pt,inner sep=0.1cm,fit=(sp1) (sp2) (sp3) (sp4) (sp5) (left) (right)] (sp) {};
          \node[font=\scriptsize,align=center,anchor=north east] () at (sp.north east) {Prediction};
        \end{pgfonlayer}

        \node[ground truth prob,minimum height=0.1cm,anchor=south] (gp3) at ([yshift=1cm]sp.north) {};
        \node[ground truth prob,minimum height=1cm,anchor=south east] (gp2) at ([xshift=-0.1cm]gp3.south west) {};
        \node[ground truth prob,minimum height=0.1cm,anchor=south east] (gp1) at ([xshift=-0.1cm]gp2.south west) {};
        \node[ground truth prob,minimum height=0.1cm,anchor=south west] (gp4) at ([xshift=0.1cm]gp3.south east) {};
        \node[ground truth prob,minimum height=0.1cm,anchor=south west] (gp5) at ([xshift=0.1cm]gp4.south east) {};

        \begin{pgfonlayer}{background}
          \coordinate (left) at ([xshift=-0.1cm]gp1.west);
          \coordinate (right) at ([xshift=0.1cm]gp5.east);
          \node[draw,densely dotted,rounded corners=5pt,inner sep=0.1cm,fit=(gp1) (gp2) (gp3) (gp4) (gp5) (left) (right)] (gp) {};
          \node[font=\scriptsize,align=center,anchor=north east] () at (gp.north east) {Ground\\Truth};
        \end{pgfonlayer}

        \coordinate (pgmid) at ([xshift=1.8cm,yshift=0.5cm]tw2.east);
        \node[pgnode] (pg1) at ([shift={(-0.3cm,0.3cm)}]pgmid) {};
        \node[pgnode] (pg2) at ([shift={(-0.3cm,-0.3cm)}]pgmid) {};
        \node[pgnode] (pg3) at ([shift={(0.3cm,0.3cm)}]pgmid) {};
        \node[pgnode] (pg4) at ([shift={(0.3cm,-0.3cm)}]pgmid) {};

        \draw[-latex,lyyblue] (pg1) to (pg3);
        \draw[-latex,lyyblue] (pg2) to (pg3);
        \draw[-latex,lyyblue] (pg1) to (pg4);
        \draw[-latex,lyyblue] (pg2) to (pg4);

        \begin{pgfonlayer}{background}
          \node[draw,rounded corners=5pt,inner sep=0.2cm,fit=(pg1) (pg2) (pg3) (pg4),label={[font=\footnotesize,align=center,label distance=3pt]90:Parameter\\Generator}] (pg) {};
        \end{pgfonlayer}

        \draw[-latex',red] (tp.east) .. controls +(east:2cm) and +(west:2cm) .. (sp.west);
        \draw[-latex',red] (gp.south) to (sp.north);

        \draw[-latex',red] (tw1.east) .. controls +(east:1cm) and +(west:1cm) .. (pg.west);
        \draw[-latex',red] (tw2.east) .. controls +(east:0.8cm) and +(west:0.8cm) .. (pg.west);
        \draw[-latex',red] (tw3.east) .. controls +(east:1cm) and +(west:1cm) .. (pg.west);

        \draw[-latex',red] (pg.east) .. controls +(east:1cm) and +(west:1cm) .. (sw1.west);
        \draw[-latex',red] (pg.east) .. controls +(east:1cm) and +(west:1cm) .. (sw2.west);

        \draw[-latex,densely dashed] (teacher.north) to (tp.south);
        \draw[-latex,densely dashed] ([xshift=-0.6cm]teacher.north) to ([xshift=-0.6cm]tp.south);
        \draw[-latex,densely dashed] ([xshift=0.6cm]teacher.north) to ([xshift=0.6cm]tp.south);

        \draw[-latex,densely dashed] (student.north) to (sp.south);
        \draw[-latex,densely dashed] ([xshift=-0.6cm]student.north) to ([xshift=-0.6cm]sp.south);
        \draw[-latex,densely dashed] ([xshift=0.6cm]student.north) to ([xshift=0.6cm]sp.south);
      \end{tikzpicture}
      \label{fig:wd}
  }
  \hspace*{\fill}
  \caption{A comparison of Knowledge Distillation and Weight Distillation (Solid red lines denote the knowledge transfer. $T_1$ and $S_1$ are the teacher and student weight matrices at the $1$st layer and so on. $L_t$ and $L_s$ are the numbers of layers in the teacher and student networks.).}
  \label{fig:compare}
\end{figure*}

To fully utilize the teacher network, we propose \emph{Weight Distillation} (WD) to transfer all the parameters of the teacher network to the student network, even if they have different numbers of weight matrices and (or) these weight matrices are of different shapes. We first use a parameter generator to predict the student network parameters from the teacher network parameters. After that, a fine-tuning process is performed to improve the quality of the transferred parameters. See \fig{fig:compare} for a comparison of KD and WD.

We test the WD method in a well-tuned Transformer-based machine translation system. The experiments are run on three machine translation benchmarks, including the WMT16 English-Roman (En-Ro), NIST12 Chinese-English (Zh-En), and WMT14 English-German (En-De) tasks. With a similar speedup, the student network trained by WD achieves BLEU improvements of 0.51$\sim$1.82 points over KD. With similar BLEU performance, the student network trained by WD is 1.11$\sim$1.39$\times$ faster than KD. More interestingly, it is found that WD is very effective in improving the student network when its model size is close to the teacher network. On the WMT14 En-De test data, our WD-based system achieves a strong result (a BLEU score of 30.77) but is 1.88$\times$ faster than the big teacher network.

\section{Background}

\subsection{Transformer}

In this work, we choose Transformer \cite{DBLP:conf/nips/VaswaniSPUJGKP17} for study because it is one of the state-of-the-art neural models in natural language processing. Transformer is a Seq2Seq model, which consists of an encoder and a decoder. The encoder maps an input sequence to a sequence of continuous representations and the decoder maps these representations to an output sequence. Both the encoder and the decoder are composed of an embedding layer and multiple hidden layers. The decoder has an additional output layer at the end.

The hidden layer in the encoder consists of a self-attention sub-layer and a feed-forward network (FFN) sub-layer. The decoder has an additional encoder-decoder attention sub-layer between the self-attention and the FFN sub-layers. For more details, we refer the reader to \cite{DBLP:conf/nips/VaswaniSPUJGKP17}.

\subsection{Knowledge Distillation}

KD encourages the student network to produce outputs close to the outputs of the teacher network. KD achieves this by:
\begin{equation}
  \bar{\mathcal{S}} = \argmin_{\mathcal{S}} \mathcal{L}(y_\mathcal{T},y_\mathcal{S})
  \label{eqn:kd}
\end{equation}
where $\mathcal{L}$ is the cross-entropy loss, $y_\mathcal{T}$ is the teacher prediction, $\mathcal{T}$ is the teacher parameters, $y_\mathcal{S}$ is the student prediction and $\mathcal{S}$ is the student parameters. In practice, \eqn{eqn:kd} serves as a regularization term.

A more effective KD variant for Seq2Seq models is proposed by \citet{DBLP:conf/emnlp/KimR16}. They replace the predicted distributions $y_\mathcal{T}$ by the generated sequences from the teacher network.

\begin{figure*}
  \centering
  \begin{tikzpicture}
    \newcommand\cube[5][]{
    \begin{scope}[#1]
      \def\x{#2}
      \def\y{#3}
      \def\z{#4}
      \path[fill=#5,draw=black] (0,0,\z) -- (\x,0,\z) -- (\x,\y,\z) -- (0,\y,\z) -- (0,0,\z); 
      \path[fill=#5,draw=black] (0,\y,\z) -- (0,\y,0) -- (\x,\y,0) -- (\x,\y,\z) -- (0,\y,\z); 
      \path[fill=#5,draw=black] (\x,0,\z) -- (\x,0,0) -- (\x,\y,0) --(\x,\y,\z) -- (\x,0,\z); 
    \end{scope}}

    \newcommand\slantrectangle[4][]{
    \begin{scope}[#1]
      \def\x{#2}
      \def\z{#3}
      \path[fill=#4,draw=black] (0,0,\z) -- (0,0,0) -- (\x,0,0) -- (\x,0,\z) -- (0,0,\z); 
    \end{scope}}

    \newcommand\rectangle[4][]{
    \begin{scope}[#1]
      \def\x{#2}
      \def\y{#3}
      \path[fill=#4,draw=black] (0,0,0) -- (\x,0,0) -- (\x,\y,0) -- (0,\y,0) -- (0,0,0); 
    \end{scope}}

    \cube[local bounding box=l1]{1.5}{0.5}{1.2}{grayblue}
    \cube[local bounding box=l2,yshift=0.5cm]{1.5}{0.5}{1.2}{grayblue}
    \cube[local bounding box=l3,yshift=2*0.5cm]{1.5}{0.5}{1.2}{grayblue}
    \cube[local bounding box=l4,yshift=3*0.5cm]{1.5}{0.5}{1.2}{lightgray}
    \cube[local bounding box=l5,yshift=4*0.5cm]{1.5}{0.5}{1.2}{lightgray}
    \cube[local bounding box=l6,yshift=5*0.5cm]{1.5}{0.5}{1.2}{lightgray}
    \draw [decorate,decoration={brace}] ([xshift=-3pt]l1.south west) to node [auto,rotate=90,anchor=south,font=\scriptsize] {$L_t=6$} ([xshift=-3pt,yshift=0.5cm]l6.south west);
    \draw [decorate,decoration={brace,mirror}] ([yshift=-3pt]l1.south west) to node [auto,below,font=\scriptsize] {$O_t=2048$} ([yshift=-3pt,xshift=-0.5cm]l1.south east);
    \draw [decorate,decoration={brace,mirror}] ([xshift=-0.5cm+3pt,yshift=-3pt]l1.south east) to node [auto,below,rotate=45,anchor=north,font=\scriptsize] {$I_t=512$} ([yshift=0.5cm-3pt,xshift=3pt]l1.south east);

    \node[font=\small,anchor=south] () at ([yshift=3pt]l6.north) {$\mathcal{T}$};
    \node[font=\small,anchor=north,align=center] () at ([yshift=-0.8cm]l1.south) {Teacher\\Network};
    \draw[-latex,grayyellow,line width=3pt] ([xshift=0.6cm]l5.east) to node [auto,above,align=center,font=\small,rotate=45,text=black,pos=0.4] {Group 1} ([shift={(1.6cm,1cm)}]l5.east);
    \draw[-latex,lightblue,line width=3pt] ([xshift=0.6cm]l2.east) to node [auto,below,align=center,font=\small,rotate=-45,text=black,pos=0.4] {Group 2} ([shift={(1.6cm,-1cm)}]l2.east);

    \cube[local bounding box=l1,xshift=3.8cm,yshift=-1cm]{1.5}{0.5}{1.2}{grayblue}
    \cube[local bounding box=l2,xshift=3.8cm,yshift=-0.5cm]{1.5}{0.5}{1.2}{grayblue}
    \cube[local bounding box=l3,xshift=3.8cm,yshift=0cm]{1.5}{0.5}{1.2}{grayblue}
    \draw [decorate,decoration={brace,mirror}] ([yshift=-3pt]l1.south west) to node [auto,below,align=center,font=\scriptsize] {$O_s=2048$} ([yshift=-3pt,xshift=-0.5cm]l1.south east);
    \draw [decorate,decoration={brace,mirror}] ([xshift=-0.5cm+3pt,yshift=-3pt]l1.south east) to node [auto,below,rotate=45,anchor=north,align=center,font=\scriptsize] {$I_s=512$} ([yshift=0.5cm-3pt,xshift=3pt]l1.south east);
    \draw [decorate,decoration={brace,mirror}] ([xshift=3pt,yshift=-0.5cm]l1.north east) to node [auto,right,rotate=90,anchor=north,font=\scriptsize] {$L_t/L_s=3$} ([xshift=3pt]l3.north east);
    \draw[-latex,lightblue,line width=3pt] ([xshift=0.75cm]l2.east) -- ([xshift=1.95cm]l2.east);

    \cube[local bounding box=l4,xshift=3.8cm,yshift=2.5cm]{1.5}{0.5}{1.2}{lightgray}
    \cube[local bounding box=l5,xshift=3.8cm,yshift=3cm]{1.5}{0.5}{1.2}{lightgray}
    \cube[local bounding box=l6,xshift=3.8cm,yshift=3.5cm]{1.5}{0.5}{1.2}{lightgray}
    \draw [decorate,decoration={brace,mirror}] ([yshift=-3pt]l4.south west) to node [auto,below,align=center,font=\scriptsize] {$O_s=2048$} ([yshift=-3pt,xshift=-0.5cm]l4.south east);
    \draw [decorate,decoration={brace,mirror}] ([xshift=-0.5cm+3pt,yshift=-3pt]l4.south east) to node [auto,below,rotate=45,anchor=north,align=center,font=\scriptsize] {$I_s=512$} ([yshift=0.5cm-3pt,xshift=3pt]l4.south east);
    \draw [decorate,decoration={brace,mirror}] ([xshift=3pt,yshift=-0.5cm]l4.north east) to node [auto,right,rotate=90,anchor=north,font=\scriptsize] {$L_t/L_s=3$} ([xshift=3pt]l6.north east);
    \draw[-latex,grayyellow,line width=3pt] ([xshift=0.75cm]l5.east) -- ([xshift=1.95cm]l5.east);

    \cube[local bounding box=l1,xshift=8cm,yshift=-1cm]{1.5}{0.5}{1.2}{grayblue}
    \cube[local bounding box=l2,xshift=8cm,yshift=-0.5cm]{1.5}{0.5}{1.2}{grayblue}
    \cube[local bounding box=l3,xshift=8cm,yshift=0cm]{1.5}{0.5}{1.2}{grayblue}
    \rectangle[local bounding box=wl,xshift=10.2cm,yshift=-1cm]{0.5}{1.5}{lightblue}
    \draw [decorate,decoration={brace}] ([xshift=3pt]wl.north east) to node [auto,right,rotate=90,anchor=north,font=\scriptsize] {$L_t/L_s=3$} ([xshift=3pt]wl.south east);
    \draw [decorate,decoration={brace,mirror}] ([yshift=-3pt]wl.south west) to node [auto,below,align=center,font=\scriptsize] {$1$} ([yshift=-3pt]wl.south east);
    \slantrectangle[local bounding box=wo,xshift=7.5cm,yshift=-1.9cm]{1.5}{1.2}{lightblue}
    \slantrectangle[local bounding box=wi,xshift=10.25cm,yshift=-1.5cm]{0.5}{1.2}{lightblue}
    \draw [decorate,decoration={brace,mirror}] ([yshift=-3pt]wo.south west) to node [auto,below,align=center,font=\scriptsize] {$O_t=2048$} ([yshift=-3pt,xshift=-0.5cm]wo.south east);
    \draw [decorate,decoration={brace,mirror}] ([xshift=-0.5cm+3pt,yshift=-3pt]wo.south east) to node [auto,below,rotate=45,anchor=north,align=center,font=\scriptsize] {$O_s=$\\$1024$} ([yshift=0.5cm-3pt,xshift=3pt]wo.south east);
    \draw [decorate,decoration={brace,mirror}] ([yshift=-3pt]wi.south west) to node [auto,below,align=center,font=\scriptsize] {$I_s=$\\$256$} ([yshift=-3pt,xshift=-0.5cm]wi.south east);
    \draw [decorate,decoration={brace,mirror}] ([xshift=-0.5cm+3pt,yshift=-3pt]wi.south east) to node [auto,below,rotate=45,anchor=north,align=center,font=\scriptsize] {$I_t=$\\$512$} ([yshift=0.5cm-3pt,xshift=3pt]wi.south east);
    \node[font=\small,anchor=west,inner sep=0pt] () at ([xshift=0.25cm]l2.east) {$\times$};
    \node[font=\tiny] () at (wl.center) {$W_L$};
    \node[font=\small,anchor=north,inner sep=0pt] () at ([shift={(-0.25cm,-0.1cm)}]l1.south) {$\times$};
    \node[font=\small,anchor=west,inner sep=0pt] () at ([shift={(0.025cm,-0.5cm)}]l1.east) {$\times$};
    \node[font=\tiny] () at (wo.center) {$W^T_O$};
    \node[font=\tiny] () at (wi.center) {$W_I$};
    \draw[-latex,lightblue,line width=3pt] ([shift={(0.7cm,-0.2cm)}]wl.east) to node [auto,below,text=black,font=\small,rotate=45,anchor=north,pos=0.4] {\eqn{eqn:pg-last}} ([shift={(1.8cm,1.2cm-0.2cm)}]wl.east);

    \cube[local bounding box=l4,xshift=8cm,yshift=2.5cm]{1.5}{0.5}{1.2}{lightgray}
    \cube[local bounding box=l5,xshift=8cm,yshift=3cm]{1.5}{0.5}{1.2}{lightgray}
    \cube[local bounding box=l6,xshift=8cm,yshift=3.5cm]{1.5}{0.5}{1.2}{lightgray}
    \rectangle[local bounding box=wl,xshift=10.2cm,yshift=2.5cm]{0.5}{1.5}{grayyellow}
    \draw [decorate,decoration={brace}] ([xshift=3pt]wl.north east) to node [auto,right,rotate=90,anchor=north,font=\scriptsize] {$L_t/L_s=3$} ([xshift=3pt]wl.south east);
    \draw [decorate,decoration={brace,mirror}] ([yshift=-3pt]wl.south west) to node [auto,below,align=center,font=\scriptsize] {$1$} ([yshift=-3pt]wl.south east);
    \slantrectangle[local bounding box=wo,xshift=7.5cm,yshift=1.6cm]{1.5}{1.2}{grayyellow}
    \slantrectangle[local bounding box=wi,xshift=10.25cm,yshift=2cm]{0.5}{1.2}{grayyellow}
    \draw [decorate,decoration={brace,mirror}] ([yshift=-3pt]wo.south west) to node [auto,below,align=center,font=\scriptsize] {$O_t=2048$} ([yshift=-3pt,xshift=-0.5cm]wo.south east);
    \draw [decorate,decoration={brace,mirror}] ([xshift=-0.5cm+3pt,yshift=-3pt]wo.south east) to node [auto,below,rotate=45,anchor=north,align=center,font=\scriptsize] {$O_s=$\\$1024$} ([yshift=0.5cm-3pt,xshift=3pt]wo.south east);
    \draw [decorate,decoration={brace,mirror}] ([yshift=-3pt]wi.south west) to node [auto,below,align=center,font=\scriptsize] {$I_s=$\\$256$} ([yshift=-3pt,xshift=-0.5cm]wi.south east);
    \draw [decorate,decoration={brace,mirror}] ([xshift=-0.5cm+3pt,yshift=-3pt]wi.south east) to node [auto,below,rotate=45,anchor=north,align=center,font=\scriptsize] {$I_t=$\\$512$} ([yshift=0.5cm-3pt,xshift=3pt]wi.south east);
    \node[font=\small,anchor=west,inner sep=0pt] () at ([xshift=0.25cm]l5.east) {$\times$};
    \node[font=\tiny] () at (wl.center) {$W_L$};
    \node[font=\small,anchor=north,inner sep=0pt] () at ([shift={(-0.25cm,-0.1cm)}]l4.south) {$\times$};
    \node[font=\small,anchor=west,inner sep=0pt] () at ([shift={(0.025cm,-0.5cm)}]l4.east) {$\times$};
    \node[font=\tiny] () at (wo.center) {$W^T_O$};
    \node[font=\tiny] () at (wi.center) {$W_I$};
    \draw[-latex,grayyellow,line width=3pt] ([shift={(0.7cm,-0.2cm)}]wl.east) to node [auto,above,text=black,font=\small,rotate=-45,anchor=south,pos=0.4] {\eqn{eqn:pg-last}} ([shift={(1.8cm,-1.2cm-0.2cm)}]wl.east);

    \cube[local bounding box=l1,xshift=13cm,yshift=1cm]{0.75}{0.5}{0.5}{grayblue}
    \cube[local bounding box=l2,xshift=13cm,yshift=1.5cm]{0.75}{0.5}{0.5}{lightgray}
    \draw [decorate,decoration={brace,mirror}] ([xshift=3pt,yshift=-0.5cm]l1.north east) to node [auto,right,rotate=90,anchor=north,font=\scriptsize] {$L_s=2$} ([xshift=3pt]l2.north east);
    \draw [decorate,decoration={brace,mirror}] ([yshift=-3pt]l1.south west) to node [auto,below,align=center,font=\scriptsize] {$O_s=$\\$1024$} ([yshift=-3pt,xshift=-0.2cm]l1.south east);
    \draw [decorate,decoration={brace,mirror}] ([xshift=-0.2cm+3pt,yshift=-3pt]l1.south east) to node [auto,below,rotate=45,anchor=north,align=center,font=\scriptsize] {$I_s=$\\$256$} ([yshift=0.2cm-3pt,xshift=3pt]l1.south east);

    \node[font=\small,anchor=south] () at ([yshift=3pt]l2.north) {$\mathcal{S}$};
    \node[font=\small,anchor=north,align=center] () at ([yshift=-2.1cm]l1.south) {Student\\Network};
  \end{tikzpicture}
  \caption{A running example of the Parameter Generator. We take the transformation of $W_1$ in \eqn{eqn:ffn} from the teacher to the student as an example. The teacher (stacked large cubes in the left) contains $L_t=6$ weights ($W_1$) with each weight from different layers. $W_1$ (a single cube) in the teacher has an input dimension $I_t$ of 512 and an output dimension $O_t$ of 2048. The student (stacked small cubes in the right) contains only $L_s=2$ weights ($W_1$) with input dimension $I_s=256$ and output dimension $O_s=1024$.}
  \label{fig:pg}
\end{figure*}
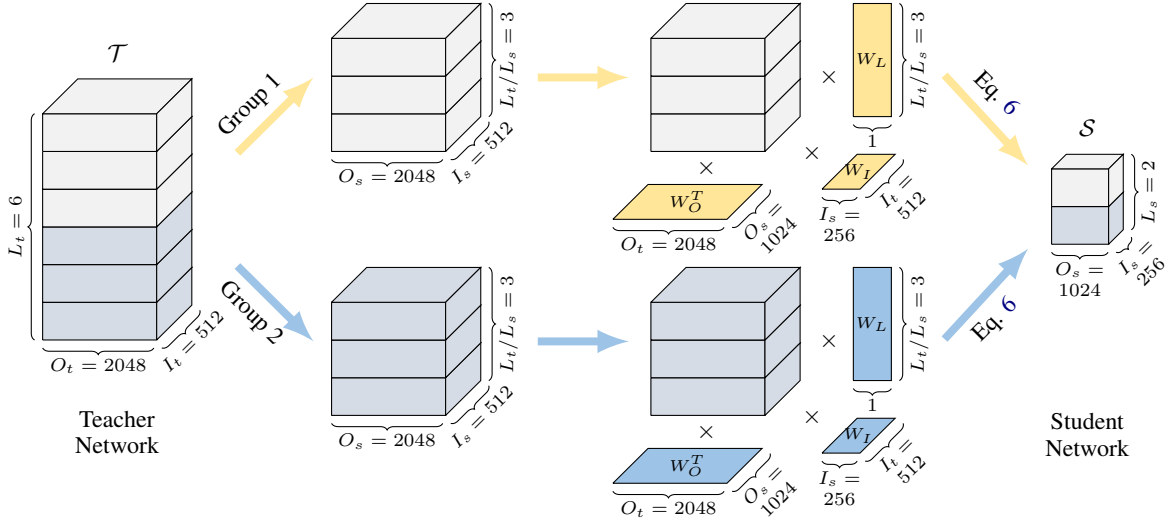

\section{Weight Distillation}

\subsection{The Parameter Generator}

The proposed parameter generator transforms the teacher parameters $\mathcal{T}$ to the student parameters $\mathcal{S}$. It is applied to the encoder and decoder separately. 

The process is simple: it first groups weight matrices in the teacher network into different subsets, and then each subset is used to generate a weight matrix in the student network. Though using all teacher weights to predict student weights is possible, its efficiency becomes an issue. For instance, the number of parameters in a simple linear transformation will be the product of the numbers of entries in its input and output, where in our case these input and output contain billions of entries (from the teacher and student weights), making it intractable to keep this simple linear transformation in the memory. Grouping is an effective way to reduce it to light-weighted transformation problems. Here we take the encoder as an example for the following discussion.

\subsubsection{Weight Grouping}

The left of \fig{fig:pg} shows an example of weight grouping for one group with two subsets.

Before the discussion, we define the weight \emph{class} as a weight matrix from the network formulation, and the weight \emph{instance} as the instantiation of a weight class. Take the FFN for an example. Its formulation is defined as:
\begin{equation}
  \mathrm{FFN}(x)=\max(xW_1+b_1,0)W_2+b_2
  \label{eqn:ffn}
\end{equation}
where $W_1$, $b_1$, $W_2$ and $b_2$ are learnable weight matrices. In this case, $W_1$ in \eqn{eqn:ffn} defines a weight class. Then all the corresponding weight matrices from FFNs in different layers of the network are the instantiations of this $W_1$ weight class.

From this sense, a weight class determines the role of its instantiations in design, e.g., extracting features for $W_1$ in \eqn{eqn:ffn}. This means that when transferring parameters, different weight classes will contribute little to each other as they have different roles. Therefore, when predicting a student weight matrix, it is sufficient to consider the teacher weight matrices with the same weight class only, which makes the prediction efficient. So our parameter generator groups the teacher weight matrices by the weight class they belong to, i.e., different weight classes clusters all their instantiations to form their own groups. In the previous example, the $W_1$ weight class will form a group $\left[T_{1},T_{2},\cdots,T_{L_t}\right]$, where each $T_i$ is the $W_1$ weight instance in the $i$-th FFN and $L_t$ is the number of layers in the teacher network. These weight matrices are then used to generate the $W_1$ weight instances in the student network.

The parameter generator further divides each group into smaller subsets with weight matrices from adjacent layers, because the adjacent layers function similarly \cite{DBLP:conf/acl/JawaharSS19} and so as their weights. This way additionally makes the later transformation more light-weighted. Namely, given a group of $L_t$ weight matrices, the parameter generator splits it into $L_s$ subsets, where $L_s$ is the number of layers in the student network. For example, the $i$-th subset of the group of $W_1$ weight class in the previous example will be $\left[T_{(i-1)*L_t/L_s+1},T_{(i-1)*L_t/L_s+2},\cdots,T_{i*L_t/L_s}\right]$. This subset is used to generate the weight matrix $S_{i}$, which corresponds to $W_1$ weight instance in the $i$-th FFN of the student network.

\subsubsection{Weight Transformation}

Given a subset of teacher weight matrices, the parameter generator then transforms them to the desired student weight matrix, as shown in the right of \fig{fig:pg}. 

Let us see the process of generating the weight matrix $S \in \mathbb{R}^{I_s \times O_s}$ from the subset $\left[T_{1},T_{2},\cdots,T_{L_t/L_s}\right]$ with each $T_{i} \in \mathbb{R}^{I_t \times O_t}$, where $I_s$ and $O_s$ are the input and output dimensions of the student weight matrix, $I_t$ and $O_t$ are the input and output dimensions of the teacher weight matrix. The parameter generator first stacks all weight matrices in this subset into a tensor $\hat{T} \in \mathbb{R}^{I_t \times O_t \times L_t/L_s}$. Then it uses three learnable weight matrices, $W_{I} \in \mathbb{R}^{I_t \times I_s},W_{O} \in \mathbb{R}^{O_t \times O_s},W_{L} \in \mathbb{R}^{L_t/L_s \times 1}$, to transform $\hat{T}$ to the shape $I_s \times O_s \times 1$ sequentially:
\begin{eqnarray}
    \hat{T}_{\cdot j k} & \leftarrow & \hat{T}_{\cdot j k}W_{I},\forall j \in [1,O_t],k \in [1,L'] \label{eqn:pg-first-i}\\
    \hat{T}_{j \cdot k} & \leftarrow & \hat{T}_{j \cdot k}W_{O},\forall j \in [1,I_s],k \in [1,L'] \label{eqn:pg-first-o}\\
    \hat{T}_{j k \cdot} & \leftarrow & \hat{T}_{j k \cdot}W_{L},\forall j \in [1,I_s],k \in [1,O_s] \label{eqn:pg-first-l}
\end{eqnarray}
\noindent where $L'=L_t/L_s$.

Finally we transform $\hat{T}$ (with $1$ in its shape get eliminated) to produce $S$, as follows:
\begin{equation}
  S = \text{tanh}(\hat{T}) \odot W + B
  \label{eqn:pg-last}
\end{equation}
\noindent where $W$ and $B$ are learnable weight matrices of the parameter generator and have the same shape as $\hat{T}$. $\odot$ denotes the Hadamard product. The tanh function provides non-linearity. $W$ and $B$ are used to scale and shift the tanh output to any desirable value. Note that we do not share $W_I$, $W_O$, $W_L$, $W$ and $B$ when generating different $S$. If the encoder is of the same size in both the teacher and student networks, only \eqn{eqn:pg-last} is needed to map each weight matrix from the teacher network to the student network.

\subsection{Training}

There are two training phases in WD: In the first phase (Phase 1), we train the parameter generator $\pi=\{W_I,W_O,W_L,W,B\}$ to predict the student network $\mathcal{S}$; In the second phase (Phase 2), we fine-tune the generated student network $\mathcal{S}$ to obtain better results. Phase 2 is necessary because the parameter generator is simply a feed-forward network with one hidden layer and thus has no enough capacity to produce a good enough student network at once. A more sophisticated parameter generator is an alternative, but it is expensive due to its large input and output spaces.

The task of Phase 1 is to minimize the loss of the student network with parameters $\mathcal{S}$ predicted by the parameter generator $\pi$ from the teacher parameters $\mathcal{T}$. The objective of Phase 1 is:
\begin{eqnarray}
  \bar{\pi} & = & \argmin_{\pi} [(1-\alpha)\mathcal{L}(y_\mathcal{T},y_\pi) + \nonumber \\
            &   &  \hspace{3.5em} \alpha \mathcal{L}(y,y_\pi)]
  \label{eqn:objective}
\end{eqnarray}

\noindent where $\mathcal{L}$ is the cross-entropy loss, $y_\mathcal{T}$ is the teacher prediction, $y_\pi$ is the prediction of the student network generated by the parameter generator $\pi$, $y$ is the ground truth, and $\alpha$ is a hyper-parameter that balances two losses and is set to 0.5 by default. The first term of \eqn{eqn:objective} is the KD loss as in \eqn{eqn:kd}, and the second term is the standard loss.

The objective of Phase 2 has the same form as \eqn{eqn:objective}, except that it optimizes $\mathcal{S}$ instead of $\pi$, like this:
\begin{eqnarray}
  \bar{\mathcal{S}} & = & \argmin_{\mathcal{S}} [(1-\alpha)\mathcal{L}(y_\mathcal{T},y_\mathcal{S}) + \nonumber \\
                    &   & \hspace{3.5em} \alpha \mathcal{L}(y,y_\mathcal{S})]
  \label{eqn:objective-phase2}
\end{eqnarray}


\begin{table*}[t!]
  \centering
  \begin{tabular}{c|l|c|c|c|c|c|r|r|c}
  \hline
  &
  \multicolumn{1}{c|}{System} &
  \multicolumn{1}{c|}{Depth} &
  \multicolumn{1}{c|}{Width} &
  \multicolumn{1}{c|}{Test} &
  \multicolumn{1}{c|}{$\mathrm{\Delta}_{\text{BLEU}}$} &
  \multicolumn{1}{c|}{Valid} &
  Params &
  \multicolumn{1}{c|}{Speed} &
  Speedup \\
  \hline
  \multirow{7}{*}{\rotatebox{90}{WMT16 En-Ro}} &
  \multirow{1}{*}{Teacher} & 6 & 512 & 31.64 & - & 32.07 & 83M & 138.35 sent./s & 1.00$\times$ \\
  \cline{2-10}
  & \multirow{1}{*}{\tinystu{}} & 1 & 256 & 29.65 & - & 29.73 & 45M & 323.26 sent./s & 2.34$\times$ \\
  & \multirow{1}{*}{\quad + KD} & 1 & 256 & 30.03 & \ \ 0.00 & 29.98 & 45M & 347.07 sent./s & 2.51$\times$ \\
  & \multirow{1}{*}{\quad + WD} & 1 & 256 & 30.89 & +0.86 & 30.89 & 45M & 359.53 sent./s & 2.60$\times$ \\
  \cline{2-10}
  & \multirow{1}{*}{\smallstu{}} & 2 & 512 & 31.22 & - & 31.19 & 66M & 281.31 sent./s & 2.03$\times$ \\
  & \multirow{1}{*}{\quad + KD} & 2 & 512 & 30.97 & \ \ 0.00 & 30.77 & 66M & 289.11 sent./s & 2.09$\times$ \\
  & \multirow{1}{*}{\quad + WD} & 2 & 512 & 31.65 & +0.68 & 31.27 & 66M & 289.80 sent./s & 2.09$\times$ \\
  \hline
  \multirow{7}{*}{\rotatebox{90}{NIST12 Zh-En}} &
  \multirow{1}{*}{Teacher} & 6 & 512 & 45.14 & - & 51.91 & 102M & 88.42 sent./s & 1.00$\times$ \\
  \cline{2-10}
  & \multirow{1}{*}{\tinystu{}} & 1 & 256 & 41.90 & - & 48.28 & 60M & 225.46 sent./s & 2.55$\times$ \\
  & \multirow{1}{*}{\quad + KD} & 1 & 256 & 42.78 & \ \ 0.00 & 49.71 & 60M & 214.06 sent./s & 2.42$\times$ \\
  & \multirow{1}{*}{\quad + WD} & 1 & 256 & 44.60 & +1.82 & 51.56 & 60M & 247.90 sent./s & 2.80$\times$ \\
  \cline{2-10}
  & \multirow{1}{*}{\smallstu{}} & 2 & 512 & 44.30 & - & 50.83 & 85M & 194.23 sent./s & 2.20$\times$ \\
  & \multirow{1}{*}{\quad + KD} & 2 & 512 & 44.89 & \ \ 0.00 & 51.87 & 85M & 199.74 sent./s & 2.26$\times$ \\
  & \multirow{1}{*}{\quad + WD} & 2 & 512 & 46.20 & +1.31 & 53.04 & 85M & 199.29 sent./s & 2.25$\times$ \\
  \hline
  \multirow{7}{*}{\rotatebox{90}{WMT14 En-De}} &
  \multirow{1}{*}{Teacher} & 6 & 512 & 27.47 & - & 26.79 & 96M & 158.29 sent./s & 1.00$\times$ \\
  \cline{2-10}
  & \multirow{1}{*}{\tinystu{}} & 1 & 256 & 24.62 & - & 24.88 & 55M & 321.79 sent./s & 2.03$\times$ \\
  & \multirow{1}{*}{\quad + KD} & 1 & 256 & 26.51 & \ \ 0.00 & 26.01 & 55M & 412.91 sent./s & 2.61$\times$ \\
  & \multirow{1}{*}{\quad + WD} & 1 & 256 & 27.12 & +0.61 & 26.42 & 55M & 406.68 sent./s & 2.57$\times$ \\
  \cline{2-10}
  & \multirow{1}{*}{\smallstu{}} & 2 & 512 & 26.68 & - & 26.07 & 80M & 281.97 sent./s & 1.78$\times$ \\
  & \multirow{1}{*}{\quad + KD} & 2 & 512 & 27.47 & \ \ 0.00 & 26.54 & 80M & 306.91 sent./s & 1.94$\times$ \\
  & \multirow{1}{*}{\quad + WD} & 2 & 512 & 28.18 & +0.71 & 26.97 & 80M & 309.11 sent./s & 1.95$\times$ \\
  \hline
  \end{tabular}
  \caption{Results of Transformer-base on different tasks (sent./s: translated sentences per second).}
  \label{tab:main_result_base}
\end{table*}

\section{Experiments}

\subsection{Datasets}

We evaluate our methods on the WMT16 English-Roman (En-Ro), NIST12 Chinese-English (Zh-En), and WMT14 English-German (En-De) tasks.

For the En-Ro task, we use the WMT16 English-Roman dataset (610K pairs). We choose \emph{newsdev-2016} as the validation set and \emph{newstest-2016} as the test set. For the Zh-En task, we use 1.8M sentence Chinese-English bitext provided within NIST12 OpenMT\footnote{LDC2000T46, LDC2000T47, LDC2000T50, LDC2003E14, LDC2005T10, LDC2002E18, LDC2007T09, LDC2004T08}. We choose the evaluation data of \emph{mt06} as the validation set, and \emph{mt08} as the test set. For the En-De task, we use the WMT14 English-German dataset (4.5M pairs). We share the source and target vocabularies. We choose \emph{newstest-2013} as the validation set and \emph{newstest-2014} as the test set.

For all datasets, we tokenize every sentence using the script in the Moses toolkit and segment every word into subword units using Byte-Pair Encoding \cite{DBLP:conf/acl/SennrichHB16a}. The number of the BPE merge operations is set to 32K. We remove sentences with more than 250 subword units \cite{DBLP:conf/acl/XiaoZZL12}. In addition, we evaluate the results using \texttt{multi-bleu.perl}.

\subsection{Model Setup}

Our baseline system is based on the open-source implementation of the Transformer model presented in \citet{DBLP:conf/naacl/OttEBFGNGA19}'s work.
For all machine translation tasks, we experiment with the Transformer-base (base) setting. We additionally run the Transformer-big (big) \cite{DBLP:conf/nips/VaswaniSPUJGKP17} and Transformer-deep (deep) \cite{DBLP:conf/acl/WangLXZLWC19,DBLP:conf/wmt/ZhangWCWSZRZZWM20} settings on the large En-De dataset. All systems consist of a 6-layer encoder and a 6-layer decoder, except that the Transformer-deep encoder has 48 layers (depth) \cite{DBLP:journals/corr/abs-2012-13866}. The embedding size (width) is set to 512 for Transformer-base/deep and 1,024 for Transformer-big. The FFN hidden size equals to 4$\times$ embedding size in all settings. We stop training until the model stops improving on the validation set. All experiments are done on 8 NVIDIA TITIAN V GPUs with mixed-precision training \cite{DBLP:conf/iclr/MicikeviciusNAD18}. At test time, the model is decoded with a beam of width 4/6/4, a length normalization weight of 1.0/1.0/0.6 and a batch size of 64 for the En-Ro/Zh-En/En-De tasks with half-precision.

Note that our method can also be seen as an advanced version of Tucker Decomposition \cite{tucker1966some}. So we also implement a baseline based on Tucker Decomposition. Unfortunately, this model does not converge to a good optima and performs extremely poor.

\begin{table*}[t!]
  \centering
  \begin{tabular}{c|l|c|c|c|c|c|r|r|c}
  \hline
  &
  \multicolumn{1}{c|}{System} &
  \multicolumn{1}{c|}{Depth} &
  \multicolumn{1}{c|}{Width} &
  \multicolumn{1}{c|}{Test} &
  \multicolumn{1}{c|}{$\mathrm{\Delta}_{\text{BLEU}}$} &
  \multicolumn{1}{c|}{Valid} &
  Params &
  \multicolumn{1}{c|}{Speed} &
  Speedup \\
  \hline
  \multirow{7}{*}{\rotatebox{90}{big}} &
  \multirow{1}{*}{Teacher} & 6 & 1024 & 29.11 & - & 27.66 & 281M & 123.92 sent./s & 1.00$\times$ \\
  \cline{2-10}
  & \multirow{1}{*}{\tinystu{}} & 1 & 512 & 25.83 & - & 25.33 & 150M & 353.42 sent./s & 2.85$\times$ \\
  & \multirow{1}{*}{\quad + KD} & 1 & 512 & 27.70 & \ \ 0.00 & 26.52 & 150M & 353.82 sent./s & 2.86$\times$ \\
  & \multirow{1}{*}{\quad + WD} & 1 & 512 & 28.60 & +0.90 & 26.83 & 150M & 364.67 sent./s & 2.94$\times$ \\
  \cline{2-10}
  & \multirow{1}{*}{\smallstu{}} & 2 & 1024 & 27.62 & - & 26.78 & 214M & 252.46 sent./s & 2.04$\times$ \\
  & \multirow{1}{*}{\quad + KD} & 2 & 1024 & 29.01 & \ \ 0.00 & 27.54 & 214M & 261.78 sent./s & 2.11$\times$ \\
  & \multirow{1}{*}{\quad + WD} & 2 & 1024 & 29.52 & +0.51 & 27.97 & 214M & 260.34 sent./s & 2.10$\times$ \\
  \hline
  \multirow{7}{*}{\rotatebox{90}{deep}} &
  \multirow{1}{*}{Teacher} & 6 & 512 & 29.43 & - & 27.82 & 229M & 134.26 sent./s & 1.00$\times$ \\
  \cline{2-10}
  & \multirow{1}{*}{\tinystu{}} & 1 & 256 & 26.34 & - & 26.05 & 187M & 270.30 sent./s & 2.01$\times$ \\
  & \multirow{1}{*}{\quad + KD} & 1 & 256 & 29.36 & \ \ 0.00 & 27.39 & 187M & 308.57 sent./s & 2.30$\times$ \\
  & \multirow{1}{*}{\quad + WD} & 1 & 256 & 29.92 & +0.56 & 27.99 & 187M & 285.43 sent./s & 2.13$\times$ \\
  \cline{2-10}
  & \multirow{1}{*}{\smallstu{}} & 2 & 512 & 28.06 & - & 26.51 & 212M & 245.82 sent./s & 1.83$\times$ \\
  & \multirow{1}{*}{\quad + KD} & 2 & 512 & 29.83 & \ \ 0.00 & 28.02 & 212M & 258.45 sent./s & 1.92$\times$ \\
  & \multirow{1}{*}{\quad + WD} & 2 & 512 & 30.77 & +0.94 & 28.33 & 212M & 252.69 sent./s & 1.88$\times$ \\
  \hline
  \end{tabular}
  \caption{Results of Transformer-big/deep on WMT14 En-De (sent./s: translated sentences per second).}
  \label{tab:main_result_big}
\end{table*}

For the KD baseline, we adopt \citet{DBLP:conf/emnlp/KimR16}'s method, which has proven to be the most effective for Seq2Seq models \cite{DBLP:conf/emnlp/KimJHAHGB19}. It generates the pseudo data from the source side of the bilingual corpus. The choices of student networks are based on the observation that the encoder has a greater impact on performance and the decoder dominates the decoding time \cite{DBLP:journals/corr/abs-2006-10369}. Therefore we vary the depth and width of the decoder. We test two student network configurations: \tinystu{} halves the decoder width and uses a 1-layer decoder (the fastest WD student network with the performance close to the teacher network); \smallstu{} uses a 2-layer decoder whose width is the same as the teacher network (the fastest KD student network with the performance close to the teacher network).

All hyper-parameters of WD are identical to the baseline system, except that WD uses 1/4 warmup steps in Phase 2. For the parameter generator initialization, we use \citet{DBLP:journals/jmlr/GlorotB10}'s method to initialize $W_I,W_O,W_L$ in Eqs. \ref{eqn:pg-first-i} - \ref{eqn:pg-first-l}. $W$ and $B$ in \eqn{eqn:pg-last} are initialized to constants 1 and 0 respectively. All results are the average of three identical runs with different random seeds.

\begin{table}[!t]
  \centering
  \setlength{\tabcolsep}{1.3mm}{
  \begin{tabular}{l|c|c|c|c}
    \hline
    \multicolumn{1}{c|}{\multirow{1}{*}{System}}
    &
    \multicolumn{1}{c|}{Test} &
    \multicolumn{1}{c|}{$\mathrm{\Delta}_{\text{BLEU}}$} &
    \multicolumn{1}{c|}{Valid} &
    \multicolumn{1}{c}{$\mathrm{\Delta}_{\text{BLEU}}$} \\
    \hline
    \multirow{1}{*}{\tinystu{} (KD)} & 42.78 & \ \ 0.00 & 49.71 & \ \ 0.00 \\
    \multirow{1}{*}{\quad + Init} & 43.36 & +0.58 & 50.32 & +0.61 \\
    \multirow{1}{*}{\quad + WD} & 44.60 & +1.82 & 51.56 & +1.85 \\
    \hline
    \multirow{1}{*}{\smallstu{} (KD)} & 44.89 & \ \ 0.00 & 51.87 & \ \ 0.00 \\
    \multirow{1}{*}{\quad + Init} & 45.66 & +0.77 & 52.57 & +0.70 \\
    \multirow{1}{*}{\quad + WD} & 46.20 & +1.31 & 53.04 & +1.17 \\
    \hline
  \end{tabular}
  \caption{Initialization study (Init: initialize the student network with the teacher parameters).}
  \label{tab:init}
  }
\end{table}

\subsection{Results}

\tab{tab:main_result_base} shows the results of different approaches on different student networks with Transformer-base as the teacher network. In all three tasks and different sized student networks, WD outperforms KD by 0.77, 1.57, and 0.66 BLEU points on En-Ro, Zh-En, and En-De on average. Our method (\tinystu{}) can obtain similar performance to the teacher network with only half of its parameters and is 2.57$\sim$2.80$\times$ faster, while KD (\smallstu{}) uses more parameters and has only a 1.94$\sim$2.26$\times$ speedup in the same case. We attribute the success of WD to that the parameter generator uses parameters of the teacher network to provide a good initialization for the student network, as Phase 1 behaves like the initialization, and the effectiveness of a good initialization has been widely proven \cite{DBLP:journals/jmlr/ErhanCBV10,DBLP:journals/corr/MishkinM15}. Interestingly, both KD and WD surpass the teacher network when the student network size is close to the teacher network (\smallstu{}). This is due to that KD has a form similar to data augmentation \cite{DBLP:journals/corr/abs-1912-03334}.

\tab{tab:main_result_big} shows the results of larger networks, i.e., Transformer-big/deep. The phenomenon here is similar to that in \tab{tab:main_result_base}. The acceleration on Transformer-big is more obvious than on Transformer-base (2.94$\times$ vs. 2.57$\times$ for \tinystu{} and 2.10$\times$ vs. 1.95$\times$ for \smallstu{} in WD). This is because the decoder in Transformer-big occupies a larger portion of the decoding time than in Transformer-base. But the acceleration on Transformer-deep is less obvious than on Transformer-base (2.13$\times$ vs. 2.57$\times$ for \tinystu{} and 1.88$\times$ vs. 1.95$\times$ for \smallstu{} in WD), as a deeper encoder consumes more inference time. Moreover, compared with such a strong Transformer-deep teacher, WD (\smallstu{}) can still outperform it by 1.34 BLEU points with a 1.88$\times$ speedup, achieving the state-of-the-art.

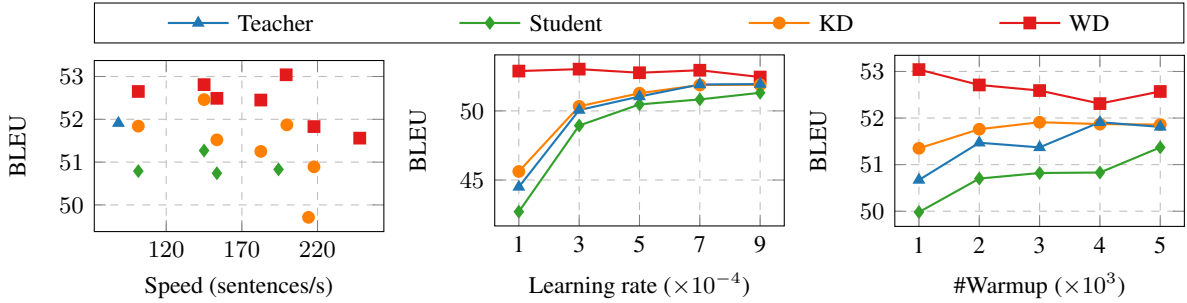
\begin{figure*}[!t]
  \hspace*{1.24cm}
  \tikz {
    \small
    \legendmargin=0.93cm
    \legendwidth=0.8cm
    \legendsep=1.8cm
    \coordinate (start) at (0,0);
    \draw[lyyblue,thick,postaction={decorate},decoration={markings,mark=at position 0.5 with {\pgfuseplotmark{triangle*}}}] ([xshift=\legendmargin]start.east) -- +(\legendwidth,0) node[black,right] (l1) {Teacher};
    \draw[lyygreen,thick,postaction={decorate},decoration={markings,mark=at position 0.5 with {\pgfuseplotmark{diamond*}}}] ([xshift=\legendsep]l1.east) -- +(\legendwidth,0) node[black,right] (l2) {Student};
    \draw[orange,thick,postaction={decorate},decoration={markings,mark=at position 0.5 with {\pgfuseplotmark{*}}}] ([xshift=\legendsep]l2.east) -- +(\legendwidth,0) node[black,right] (l3) {KD};
    \draw[lyyred,thick,postaction={decorate},decoration={markings,mark=at position 0.5 with {\pgfuseplotmark{square*}}}] ([xshift=\legendsep]l3.east) -- +(\legendwidth,0) node[black,right] (l4) {WD};
    \coordinate (end) at ([xshift=\legendmargin-3pt]l4.east);
    \begin{pgfonlayer}{background}
    \node[rectangle,draw,inner sep=1pt] [fit = (start) (l1) (l2) (l3) (l4) (end)] {};
    \end{pgfonlayer}
  }
  \\[3pt]
  \centering
  \hspace*{\fill}
  \begin{tikzpicture}
    \begin{axis}[
      width=0.7\columnwidth,height=0.5\columnwidth,
      yticklabel style={/pgf/number format/fixed,/pgf/number format/precision=1},
      ylabel={BLEU},
      ylabel near ticks,
      xlabel={Speed (sentences/s)},
      xlabel near ticks,
      enlargelimits=0.1,
      xmajorgrids=true,
      ymajorgrids=true,
      grid style=dashed,
      xtick={120,170,220,270},
      every tick label/.append style={font=\small},
      label style={font=\small},
      ylabel style={yshift=5pt},
    ]
      \addplot [
        scatter,
        only marks,
        point meta=explicit symbolic,
        scatter/classes={
          a={mark=square*,thick,lyyred},
          b={mark=*,thick,orange},
          c={mark=diamond*,thick,lyygreen},
          d={mark=triangle*,thick,lyyblue}
        }
      ]table [meta=label] {
        x y label
        101.56 52.65 a
        145.04 52.81 a
        153.50 52.49 a
        199.29 53.04 a
        182.53 52.45 a
        217.66 51.83 a
        247.90 51.56 a
        101.56 51.84 b
        145.04 52.46 b
        153.50 51.52 b
        199.74 51.87 b
        182.53 51.25 b
        217.66 50.89 b
        214.06 49.71 b
        101.56 50.79 c
        145.04 51.27 c
        153.50 50.74 c
        194.23 50.83 c
        88.42 51.91 d
      };
    \end{axis}
  \end{tikzpicture}
  \hfill
  \begin{tikzpicture}
    \begin{axis}[
      width=0.7\columnwidth,height=0.5\columnwidth,
      yticklabel style={/pgf/number format/fixed,/pgf/number format/precision=1},
      ylabel={BLEU},
      ylabel near ticks,
      xlabel={Learning rate ($\times 10^{-4}$)},
      xlabel near ticks,
      enlargelimits=0.1,
      symbolic x coords={1,3,5,7,9},
      xmajorgrids=true,
      ymajorgrids=true,
      grid style=dashed,
      xtick=data,
      every tick label/.append style={font=\small},
      label style={font=\small},
      ylabel style={yshift=5pt},
    ]
      \addplot [lyyred,thick,mark=square*] coordinates {
        (1,52.88) (3,53.02) (5,52.76) (7,52.94) (9,52.45)
      };
      \addplot [orange,thick,mark=*] coordinates {
        (1,45.62) (3,50.33) (5,51.28) (7,51.87) (9,51.89)
      };
      \addplot [lyygreen,thick,mark=diamond*] coordinates {
        (1,42.71) (3,48.94) (5,50.47) (7,50.83) (9,51.31)
      };
      \addplot [lyyblue,thick,thick,mark=triangle*] coordinates {
        (1,44.50) (3,50.06) (5,51.04) (7,51.91) (9,51.94)
      };
    \end{axis}
  \end{tikzpicture}
  \hfill
  \begin{tikzpicture}
    \begin{axis}[
      width=0.7\columnwidth,height=0.5\columnwidth,
      yticklabel style={/pgf/number format/fixed,/pgf/number format/precision=1},
      ylabel={BLEU},
      ylabel near ticks,
      xlabel={\#Warmup ($\times 10^3$)},
      xlabel near ticks,
      enlargelimits=0.1,
      symbolic x coords={1,2,3,4,5},
      xmajorgrids=true,
      ymajorgrids=true,
      grid style=dashed,
      xtick=data,
      every tick label/.append style={font=\small},
      label style={font=\small},
      ylabel style={yshift=5pt},
    ]
      \addplot [lyyred,thick,mark=square*] coordinates {
        (1,53.04) (2,52.71) (3,52.59) (4,52.31) (5,52.57)
      };
      \addplot [orange,thick,mark=*] coordinates {
        (1,51.35) (2,51.76) (3,51.91) (4,51.87) (5,51.86)
      };
      \addplot [lyygreen,thick,mark=diamond*] coordinates {
        (1,49.98) (2,50.70) (3,50.82) (4,50.83) (5,51.37)
      };
      \addplot [lyyblue,thick,thick,mark=triangle*] coordinates {
        (1,50.67) (2,51.47) (3,51.37) (4,51.91) (5,51.81)
      };
    \end{axis}
  \end{tikzpicture}
  \hspace*{\fill}
  \caption{Sensitivity analysis on \smallstu{}.}
  \label{fig:sensitive}
\end{figure*}

\begin{table*}[!t]
  \begin{minipage}[!t]{\columnwidth}
    \centering
    \setlength{\tabcolsep}{0.8mm}{
      \begin{tabular}{l|c|c|c|c}
        \hline
        \multicolumn{1}{c|}{\multirow{1}{*}{System}}
        &
        \multicolumn{1}{c|}{Test} &
        \multicolumn{1}{c|}{$\mathrm{\Delta}_{\text{BLEU}}$} &
        \multicolumn{1}{c|}{Valid} &
        \multicolumn{1}{c}{$\mathrm{\Delta}_{\text{BLEU}}$} \\
        \hline
        \multirow{1}{*}{\smallstu{}} & 44.30 & \ \ 0.00 & 50.83 & \ \ 0.00 \\
        \hline
        \multirow{1}{*}{\quad + KD} & 44.89 & +0.59 & 51.87 & +1.04 \\
        \multirow{1}{*}{\quad + Encoder} & 45.40 & +1.10 & 51.62 & +0.79 \\
        \multirow{1}{*}{\quad + Decoder} & 45.26 & +0.96 & 51.34 & +0.51 \\
        \multirow{1}{*}{\quad + Embed (Enc)} & 44.67 & +0.37 & 51.22 & +0.39 \\
        \multirow{1}{*}{\quad + Embed (Dec)} & 45.06 & +0.76 & 51.26 & +0.43 \\
        \multirow{1}{*}{\quad + Output} & 45.10 & +0.80 & 51.28 & +0.45 \\
        \hline
      \end{tabular}
      \caption{Ablation study of using different weight matrices solely.}
      \label{tab:ablation}
    }
  \end{minipage}
  \hfill
  \begin{minipage}[!t]{\columnwidth}
    \centering
    \setlength{\tabcolsep}{0.5mm}{
      \begin{tabular}{c|c|r|c|r}
        \hline
        \multicolumn{1}{c|}{\multirow{2}{*}{\diagbox{D}{W}}} &
        \multicolumn{2}{c|}{256} &
        \multicolumn{2}{c}{512} \\
        \cline{2-5}
        &
        \multicolumn{1}{c|}{BLEU$_{\text{KD/WD}}$} &
        \multicolumn{1}{c|}{Params} &
        \multicolumn{1}{c|}{BLEU$_{\text{KD/WD}}$} &
        \multicolumn{1}{c}{Params} \\
        \hline
        1 & 38.46/40.34 & 30M & 43.51/45.39 & 65M \\
        2 & 45.33/47.21 & 32M & 50.02/50.45 & 72M \\
        3 & 47.30/49.09 & 34M & 51.18/51.99 & 80M \\
        4 & 47.90/50.08 & 36M & 51.05/52.05 & 87M \\
        5 & 48.87/50.70 & 38M & 52.15/52.00 & 94M \\
        6 & 49.78/50.73 & 40M & 52.40/53.09 & 102M \\
        \hline
      \end{tabular}
      \caption{Compression study with various depth (\emph{D}) and width (\emph{W}) of both the encoder and decoder.}
      \label{tab:shape_encoder_decoder}
    }
  \end{minipage}
\end{table*}

\section{Analysis}

To better understand WD, we conduct a series of experiments on the NIST12 Zh-En validation set with the Transformer-base teacher.

\subsection{Initialization Study}
\label{sub-sec:initialization}


To test whether KD misses knowledge in parameters, we initialize the student network with the teacher parameters. If the teacher and student networks have different depths, we initialize the student network with the bottom layers of the teacher network \cite{DBLP:journals/corr/abs-1910-01108}. If they have different widths, we slice the teacher weight matrices to fit the student network \cite{wang2020hat}. \tab{tab:init} shows that initializing the student networks with the teacher parameters improves KD, supporting our claim that knowledge in parameters is complementary to KD but missed. We also see that WD outperforms this simple initialization, which implies that using all teacher parameters helps to obtain a better student.

\subsection{Sensitivity Analysis}

The left part of \fig{fig:sensitive} studies how sensitive the performance (BLEU) of different methods are to various levels of inference speedup (obtained by varying decoder depth and width). It shows that WD distributes on the upper right of the figure, which means that WD produces student networks that are consistently faster and better.

We also investigate how sensitive different methods are to the training hyper-parameters, i.e., the learning rate and warmup steps. Here we focus on Phase 2 of WD, as it directly impacts the final performance. The middle part of \fig{fig:sensitive} shows that WD can endure learning rates in a wide range, because its performance does not vary much. However, a very large learning rate still negatively impacts the performance. The right part of \fig{fig:sensitive} is the opposite, where WD is more sensitive to the warmup steps than the learning rate. This is because more warmup steps will run the network with a high learning rate in a longer period. A high learning rate has been proven to be harmful as shown in the middle part of \fig{fig:sensitive}.

\subsection{Ablation Study}

\tab{tab:ablation} studies which weight matrices in the teacher network are the most effective. It is achieved by training the parameter generator with only the intended weight matrices and without the KD loss term in \eqn{eqn:objective}. We see that using any weight matrix brings a significant improvement over the baseline. This observation shows that weight matrices in the teacher network do contain abundant knowledge. Among these, the encoder weight matrices produce the most significant result, which agrees with the previous study claiming that the encoder is more important than the decoder \cite{DBLP:conf/acl/WangLXZLWC19,bapna-etal-2018-training}.

\subsection{Compression Study}

As the previous experiments focus on a lightweight decoder for acceleration, the compression is limited as the encoder remains large. To examine the effectiveness of WD on model compression, we shrink the depth and width of the encoder and decoder simultaneously. As shown in \tab{tab:shape_encoder_decoder}, WD consistently outperforms KD by about 1 BLEU point under various compression ratios (ranging from 1.00$\times$ to 3.40$\times$). Note that decreasing the width brings more significant compression. This is because a large portion of the parameters is from the embedding matrices and the output projection. The sizes of these matrices are determined by the width and a fixed vocabulary size.

\subsection{Training Efficiency}

\begin{figure}[!t]
  \centering
  \begin{tikzpicture}
    \begin{axis}[
        width=0.7\linewidth,height=0.7\linewidth,
        view={0}{90},
        enlargelimits=false,
        ymin=-0.5,ymax=9.5,
        xmin=-0.5,xmax=9.5,
        xlabel={\#Epoch (Phase 2)},
        ylabel={\#Epoch (Phase 1)},
        ytick={0,1,2,3,4,5,6,7,8,9},
        yticklabels={1,2,3,4,5,6,7,8,9,10},
        xtick={0,1,2,3,4,5,6,7,8,9},
        xticklabels={1,2,3,4,5,6,7,8,9,10},
        label style={font=\small},
        every tick label/.append style={font=\small},
        xticklabel pos=upper,
        xtick style={draw=none},
        ytick style={draw=none},
        colormap={redyellow}{color=(yellow) color=(red)},
        colorbar,
        colorbar style={
            font=\small,
            title style={align=center,at={(0.5,1.05)},anchor=south},
            title={BLEU},
            yticklabel style={
                /pgf/number format/precision=1,
                /pgf/number format/.cd,
                    fixed,
                    fixed zerofill
            },
            at={(1.1,1)},anchor=north west,
        },
    ]
      \addplot3[matrix plot] table [meta=value] {
        x y value
        0 0 49.96
        0 1 50.82
        0 2 51.68
        0 3 51.58
        0 4 51.50
        0 5 52.09
        0 6 52.02
        0 7 51.80
        0 8 51.99
        0 9 52.25

        1 0 50.80
        1 1 51.06
        1 2 52.00
        1 3 51.85
        1 4 51.67
        1 5 51.99
        1 6 51.88
        1 7 52.17
        1 8 52.35
        1 9 52.27

        2 0 50.96
        2 1 51.36
        2 2 51.87
        2 3 51.76
        2 4 52.07
        2 5 52.42
        2 6 52.35
        2 7 52.28
        2 8 52.33
        2 9 52.26

        3 0 50.79
        3 1 51.16
        3 2 51.65
        3 3 51.73
        3 4 52.17
        3 5 51.97
        3 6 51.77
        3 7 52.53
        3 8 52.33
        3 9 52.51

        4 0 51.14
        4 1 51.64
        4 2 52.18
        4 3 52.16
        4 4 52.01
        4 5 52.44
        4 6 52.09
        4 7 52.29
        4 8 52.16
        4 9 52.53

        5 0 51.05
        5 1 51.77
        5 2 51.98
        5 3 52.45
        5 4 52.13
        5 5 52.26
        5 6 52.15
        5 7 52.39
        5 8 52.38
        5 9 52.71

        6 0 51.21
        6 1 51.96
        6 2 52.37
        6 3 52.19
        6 4 52.19
        6 5 52.64
        6 6 52.32
        6 7 52.71
        6 8 52.61
        6 9 52.71

        7 0 51.06
        7 1 51.51
        7 2 52.12
        7 3 51.87
        7 4 52.26
        7 5 52.31
        7 6 52.28
        7 7 52.33
        7 8 52.65
        7 9 52.73

        8 0 50.71
        8 1 51.83
        8 2 51.73
        8 3 52.07
        8 4 52.11
        8 5 52.41
        8 6 52.36
        8 7 52.51
        8 8 52.34
        8 9 52.73

        9 0 51.15
        9 1 51.58
        9 2 52.20
        9 3 52.00
        9 4 52.01
        9 5 52.22
        9 6 52.01
        9 7 52.36
        9 8 52.35
        9 9 52.58
      };
    \end{axis}
  \end{tikzpicture}
  \caption{Training efficiency of WD on \smallstu{}.}
  \label{fig:efficiency}
\end{figure}
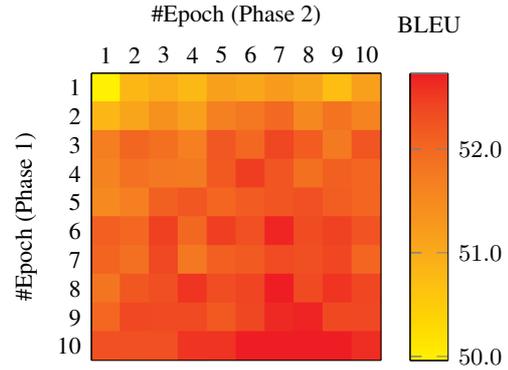

\fig{fig:efficiency} studies the training efficiency of WD by comparing the final BLEU scores when two training phases end in different epochs. As shown in \fig{fig:efficiency}, Phase 1 has little impact on Phase 2, because Phase 2 converges to optimums with similar BLEU scores once Phase 1 runs for a few epochs (say, 3 epochs). If we run Phase 1 longer, then Phase 2 converges faster. This phenomenon suggests that Phase 1 already transfers the knowledge in the teacher parameters within the first few epochs, and the remaining epochs merely do the fine-tuning (Phase 2) job. This implies that the training of WD is efficient, since we can just train the parameter generator for several epochs first, then fine-tune the generated network as in KD, and finally obtain a much better result than KD.

Though we could train the parameter generator for just a few epochs as suggested, Phase 1 is still time-consuming. The reasons are two folds: 1) the parameter generator consumes a lot of memory and we have to resort to gradient accumulation; 2) the parameter generator involves many large matrix multiplications. For the experiments in \tab{tab:main_result_base} and \tab{tab:main_result_big}, it takes us 0.66 days for WD to finish training on average, whereas 0.55 days for the teacher network baseline and 0.31 days for both the student network baseline and KD.

\section{Related Work}

\subsection{Knowledge Distillation}
Knowledge distillation \cite{DBLP:journals/corr/HintonVD15,DBLP:journals/corr/FreitagAS17} is a widely used model acceleration and compression technique \cite{DBLP:journals/corr/abs-1909-10351,DBLP:journals/corr/abs-1910-01108,DBLP:journals/corr/abs-2004-02178}. It treats the network predictions as the knowledge learned by the teacher network, since these predicted distributions contain the ranking information on similarities among categories. It then transfers this knowledge to the student network by enforcing the student network to have similar predictions. The followed work extends this idea by providing more knowledge from different sources to the student network. FitNets \cite{DBLP:journals/corr/RomeroBKCGB14} uses not only the predictions but also the intermediate representations learned by the teacher network to supervise the student network. For the Seq2Seq model, \citet{DBLP:conf/emnlp/KimR16} proposes to use the generated sequences as the sequence-level knowledge to guide the student network training. Moreover, self-knowledge distillation \cite{DBLP:conf/ranlp/HahnC19} even shows that knowledge (representations) from the student network itself can improve the performance.

Our weight distillation, on the other hand, explores a new source of knowledge and a new way to leverage this knowledge. It transfers the knowledge in parameters of the teacher network to the student network via a parameter generator. Therefore, it is orthogonal to other knowledge distillation variants.

\subsection{Transfer Learning}
Transfer learning aims at transferring knowledge from a source domain to a target domain. Based on what knowledge is transferred to the model in the target domain, transfer learning methods can be classified into three categories \cite{DBLP:journals/tkde/PanY10}: instance-based methods reuse certain parts of the data in the source domain \cite{DBLP:conf/acl/JiangZ07,DBLP:conf/icml/DaiYXY07}; feature-based methods use the representation from the model learned in the source domain as the input \cite{DBLP:conf/naacl/PetersNIGCLZ18,DBLP:conf/kdd/GaoFJH08}; parameter-based methods directly fine-tune the model learned in the source domain with the target domain data \cite{DBLP:conf/nips/YangDYCSL19,DBLP:journals/corr/abs-1907-11692,DBLP:conf/naacl/DevlinCLT19}.

Perhaps the most related work is \citet{DBLP:conf/emnlp/PlataniosSNM18}'s work. Their method falls into the parameter-based category. They use a universal parameter generator to share the knowledge among translation tasks. This parameter generator produces a translation model from a given language-specific embedding. Though we similarly employ the idea of a parameter generator, our weight distillation aims at transferring knowledge from one model to another rather than from one translation task to another. Therefore our parameter generator takes a model instead of a language-specific embedding as its input and is only used once.

\section{Conclusion}

In this work, we propose weight distillation to transfer knowledge in the parameters of the teacher network to the student network. It generates the student network from the teacher network via a parameter generator. Our experiments on three machine translation tasks show that weight distillation consistently outperforms knowledge distillation by producing a faster and better student network.

\section*{Acknowledgments}

This work was supported in part by the National Science Foundation of China (Nos. 61876035 and 61732005), the National Key R\&D Program of China (No. 2019QY1801), and the Ministry of Science and Technology of the PRC (Nos. 2019YFF0303002 and 2020AAA0107900). The authors would like to thank anonymous reviewers for their comments.

\bibliographystyle{acl_natbib}
\bibliography{acl2021}

\begin{thebibliography}{35}
\expandafter\ifx\csname natexlab\endcsname\relax\def\natexlab#1{#1}\fi

\bibitem[{Bapna et~al.(2018)Bapna, Chen, Firat, Cao, and
  Wu}]{bapna-etal-2018-training}
Ankur Bapna, Mia Chen, Orhan Firat, Yuan Cao, and Yonghui Wu. 2018.
\newblock Training deeper neural machine translation models with transparent
  attention.
\newblock In \emph{Proceedings of the 2018 Conference on Empirical Methods in
  Natural Language Processing}, pages 3028--3033, Brussels, Belgium.
  Association for Computational Linguistics.

\bibitem[{Dai et~al.(2007)Dai, Yang, Xue, and Yu}]{DBLP:conf/icml/DaiYXY07}
Wenyuan Dai, Qiang Yang, Gui{-}Rong Xue, and Yong Yu. 2007.
\newblock Boosting for transfer learning.
\newblock In \emph{Machine Learning, Proceedings of the Twenty-Fourth
  International Conference {(ICML} 2007), Corvallis, Oregon, USA, June 20-24,
  2007}, volume 227 of \emph{{ACM} International Conference Proceeding Series},
  pages 193--200. {ACM}.

\bibitem[{Devlin et~al.(2019)Devlin, Chang, Lee, and
  Toutanova}]{DBLP:conf/naacl/DevlinCLT19}
Jacob Devlin, Ming{-}Wei Chang, Kenton Lee, and Kristina Toutanova. 2019.
\newblock {BERT:} pre-training of deep bidirectional transformers for language
  understanding.
\newblock In \emph{Proceedings of the 2019 Conference of the North American
  Chapter of the Association for Computational Linguistics: Human Language
  Technologies, {NAACL-HLT} 2019, Minneapolis, MN, USA, June 2-7, 2019, Volume
  1 (Long and Short Papers)}, pages 4171--4186. Association for Computational
  Linguistics.

\bibitem[{Erhan et~al.(2010)Erhan, Courville, Bengio, and
  Vincent}]{DBLP:journals/jmlr/ErhanCBV10}
Dumitru Erhan, Aaron~C. Courville, Yoshua Bengio, and Pascal Vincent. 2010.
\newblock Why does unsupervised pre-training help deep learning?
\newblock In \emph{Proceedings of the Thirteenth International Conference on
  Artificial Intelligence and Statistics, {AISTATS} 2010, Chia Laguna Resort,
  Sardinia, Italy, May 13-15, 2010}, volume~9 of \emph{{JMLR} Proceedings},
  pages 201--208. JMLR.org.

\bibitem[{Freitag et~al.(2017)Freitag, Al{-}Onaizan, and
  Sankaran}]{DBLP:journals/corr/FreitagAS17}
Markus Freitag, Yaser Al{-}Onaizan, and Baskaran Sankaran. 2017.
\newblock Ensemble distillation for neural machine translation.
\newblock \emph{CoRR}, abs/1702.01802.

\bibitem[{Gao et~al.(2008)Gao, Fan, Jiang, and Han}]{DBLP:conf/kdd/GaoFJH08}
Jing Gao, Wei Fan, Jing Jiang, and Jiawei Han. 2008.
\newblock Knowledge transfer via multiple model local structure mapping.
\newblock In \emph{Proceedings of the 14th {ACM} {SIGKDD} International
  Conference on Knowledge Discovery and Data Mining, Las Vegas, Nevada, USA,
  August 24-27, 2008}, pages 283--291. {ACM}.

\bibitem[{Glorot and Bengio(2010)}]{DBLP:journals/jmlr/GlorotB10}
Xavier Glorot and Yoshua Bengio. 2010.
\newblock Understanding the difficulty of training deep feedforward neural
  networks.
\newblock In \emph{Proceedings of the Thirteenth International Conference on
  Artificial Intelligence and Statistics, {AISTATS} 2010, Chia Laguna Resort,
  Sardinia, Italy, May 13-15, 2010}, volume~9 of \emph{{JMLR} Proceedings},
  pages 249--256. JMLR.org.

\bibitem[{Gordon and Duh(2019)}]{DBLP:journals/corr/abs-1912-03334}
Mitchell~A. Gordon and Kevin Duh. 2019.
\newblock Explaining sequence-level knowledge distillation as data-augmentation
  for neural machine translation.
\newblock \emph{CoRR}, abs/1912.03334.

\bibitem[{Hahn and Choi(2019)}]{DBLP:conf/ranlp/HahnC19}
Sangchul Hahn and Heeyoul Choi. 2019.
\newblock Self-knowledge distillation in natural language processing.
\newblock In \emph{Proceedings of the International Conference on Recent
  Advances in Natural Language Processing, {RANLP} 2019, Varna, Bulgaria,
  September 2-4, 2019}, pages 423--430. {INCOMA} Ltd.

\bibitem[{Hinton et~al.(2015)Hinton, Vinyals, and
  Dean}]{DBLP:journals/corr/HintonVD15}
Geoffrey~E. Hinton, Oriol Vinyals, and Jeffrey Dean. 2015.
\newblock Distilling the knowledge in a neural network.
\newblock \emph{CoRR}, abs/1503.02531.

\bibitem[{Jawahar et~al.(2019)Jawahar, Sagot, and
  Seddah}]{DBLP:conf/acl/JawaharSS19}
Ganesh Jawahar, Beno{\^{\i}}t Sagot, and Djam{\'{e}} Seddah. 2019.
\newblock What does {BERT} learn about the structure of language?
\newblock In \emph{Proceedings of the 57th Conference of the Association for
  Computational Linguistics, {ACL} 2019, Florence, Italy, July 28- August 2,
  2019, Volume 1: Long Papers}, pages 3651--3657. Association for Computational
  Linguistics.

\bibitem[{Jiang and Zhai(2007)}]{DBLP:conf/acl/JiangZ07}
Jing Jiang and ChengXiang Zhai. 2007.
\newblock Instance weighting for domain adaptation in {NLP}.
\newblock In \emph{{ACL} 2007, Proceedings of the 45th Annual Meeting of the
  Association for Computational Linguistics, June 23-30, 2007, Prague, Czech
  Republic}. The Association for Computational Linguistics.

\bibitem[{Jiao et~al.(2019)Jiao, Yin, Shang, Jiang, Chen, Li, Wang, and
  Liu}]{DBLP:journals/corr/abs-1909-10351}
Xiaoqi Jiao, Yichun Yin, Lifeng Shang, Xin Jiang, Xiao Chen, Linlin Li, Fang
  Wang, and Qun Liu. 2019.
\newblock Tinybert: Distilling {BERT} for natural language understanding.
\newblock \emph{CoRR}, abs/1909.10351.

\bibitem[{Kasai et~al.(2020)Kasai, Pappas, Peng, Cross, and
  Smith}]{DBLP:journals/corr/abs-2006-10369}
Jungo Kasai, Nikolaos Pappas, Hao Peng, James Cross, and Noah~A. Smith. 2020.
\newblock Deep encoder, shallow decoder: Reevaluating the speed-quality
  tradeoff in machine translation.
\newblock \emph{CoRR}, abs/2006.10369.

\bibitem[{Kim and Rush(2016)}]{DBLP:conf/emnlp/KimR16}
Yoon Kim and Alexander~M. Rush. 2016.
\newblock Sequence-level knowledge distillation.
\newblock In \emph{Proceedings of the 2016 Conference on Empirical Methods in
  Natural Language Processing, {EMNLP} 2016, Austin, Texas, USA, November 1-4,
  2016}, pages 1317--1327. The Association for Computational Linguistics.

\bibitem[{Kim et~al.(2019)Kim, Junczys{-}Dowmunt, Hassan, Aji, Heafield,
  Grundkiewicz, and Bogoychev}]{DBLP:conf/emnlp/KimJHAHGB19}
Young~Jin Kim, Marcin Junczys{-}Dowmunt, Hany Hassan, Alham~Fikri Aji, Kenneth
  Heafield, Roman Grundkiewicz, and Nikolay Bogoychev. 2019.
\newblock From research to production and back: Ludicrously fast neural machine
  translation.
\newblock In \emph{Proceedings of the 3rd Workshop on Neural Generation and
  Translation@EMNLP-IJCNLP 2019, Hong Kong, November 4, 2019}, pages 280--288.
  Association for Computational Linguistics.

\bibitem[{Li et~al.(2020)Li, Wang, Liu, Du, Xiao, Zhang, and
  Zhu}]{DBLP:journals/corr/abs-2012-13866}
Bei Li, Ziyang Wang, Hui Liu, Quan Du, Tong Xiao, Chunliang Zhang, and Jingbo
  Zhu. 2020.
\newblock Learning light-weight translation models from deep transformer.
\newblock \emph{CoRR}, abs/2012.13866.

\bibitem[{Liu et~al.(2020)Liu, Zhou, Zhao, Wang, Deng, and
  Ju}]{DBLP:journals/corr/abs-2004-02178}
Weijie Liu, Peng Zhou, Zhe Zhao, Zhiruo Wang, Haotang Deng, and Qi~Ju. 2020.
\newblock Fastbert: a self-distilling {BERT} with adaptive inference time.
\newblock \emph{CoRR}, abs/2004.02178.

\bibitem[{Liu et~al.(2019)Liu, Ott, Goyal, Du, Joshi, Chen, Levy, Lewis,
  Zettlemoyer, and Stoyanov}]{DBLP:journals/corr/abs-1907-11692}
Yinhan Liu, Myle Ott, Naman Goyal, Jingfei Du, Mandar Joshi, Danqi Chen, Omer
  Levy, Mike Lewis, Luke Zettlemoyer, and Veselin Stoyanov. 2019.
\newblock Roberta: {A} robustly optimized {BERT} pretraining approach.
\newblock \emph{CoRR}, abs/1907.11692.

\bibitem[{Micikevicius et~al.(2018)Micikevicius, Narang, Alben, Diamos, Elsen,
  Garc{\'{\i}}a, Ginsburg, Houston, Kuchaiev, Venkatesh, and
  Wu}]{DBLP:conf/iclr/MicikeviciusNAD18}
Paulius Micikevicius, Sharan Narang, Jonah Alben, Gregory~F. Diamos, Erich
  Elsen, David Garc{\'{\i}}a, Boris Ginsburg, Michael Houston, Oleksii
  Kuchaiev, Ganesh Venkatesh, and Hao Wu. 2018.
\newblock Mixed precision training.
\newblock In \emph{6th International Conference on Learning Representations,
  {ICLR} 2018, Vancouver, BC, Canada, April 30 - May 3, 2018, Conference Track
  Proceedings}. OpenReview.net.

\bibitem[{Mishkin and Matas(2016)}]{DBLP:journals/corr/MishkinM15}
Dmytro Mishkin and Jiri Matas. 2016.
\newblock All you need is a good init.
\newblock In \emph{4th International Conference on Learning Representations,
  {ICLR} 2016, San Juan, Puerto Rico, May 2-4, 2016, Conference Track
  Proceedings}.

\bibitem[{Ott et~al.(2019)Ott, Edunov, Baevski, Fan, Gross, Ng, Grangier, and
  Auli}]{DBLP:conf/naacl/OttEBFGNGA19}
Myle Ott, Sergey Edunov, Alexei Baevski, Angela Fan, Sam Gross, Nathan Ng,
  David Grangier, and Michael Auli. 2019.
\newblock fairseq: {A} fast, extensible toolkit for sequence modeling.
\newblock In \emph{Proceedings of the 2019 Conference of the North American
  Chapter of the Association for Computational Linguistics: Human Language
  Technologies, {NAACL-HLT} 2019, Minneapolis, MN, USA, June 2-7, 2019,
  Demonstrations}, pages 48--53. Association for Computational Linguistics.

\bibitem[{Pan and Yang(2010)}]{DBLP:journals/tkde/PanY10}
Sinno~Jialin Pan and Qiang Yang. 2010.
\newblock A survey on transfer learning.
\newblock \emph{{IEEE} Trans. Knowl. Data Eng.}, 22(10):1345--1359.

\bibitem[{Peters et~al.(2018)Peters, Neumann, Iyyer, Gardner, Clark, Lee, and
  Zettlemoyer}]{DBLP:conf/naacl/PetersNIGCLZ18}
Matthew~E. Peters, Mark Neumann, Mohit Iyyer, Matt Gardner, Christopher Clark,
  Kenton Lee, and Luke Zettlemoyer. 2018.
\newblock Deep contextualized word representations.
\newblock In \emph{Proceedings of the 2018 Conference of the North American
  Chapter of the Association for Computational Linguistics: Human Language
  Technologies, {NAACL-HLT} 2018, New Orleans, Louisiana, USA, June 1-6, 2018,
  Volume 1 (Long Papers)}, pages 2227--2237. Association for Computational
  Linguistics.

\bibitem[{Platanios et~al.(2018)Platanios, Sachan, Neubig, and
  Mitchell}]{DBLP:conf/emnlp/PlataniosSNM18}
Emmanouil~Antonios Platanios, Mrinmaya Sachan, Graham Neubig, and Tom~M.
  Mitchell. 2018.
\newblock Contextual parameter generation for universal neural machine
  translation.
\newblock In \emph{Proceedings of the 2018 Conference on Empirical Methods in
  Natural Language Processing, Brussels, Belgium, October 31 - November 4,
  2018}, pages 425--435. Association for Computational Linguistics.

\bibitem[{Romero et~al.(2015)Romero, Ballas, Kahou, Chassang, Gatta, and
  Bengio}]{DBLP:journals/corr/RomeroBKCGB14}
Adriana Romero, Nicolas Ballas, Samira~Ebrahimi Kahou, Antoine Chassang, Carlo
  Gatta, and Yoshua Bengio. 2015.
\newblock Fitnets: Hints for thin deep nets.
\newblock In \emph{3rd International Conference on Learning Representations,
  {ICLR} 2015, San Diego, CA, USA, May 7-9, 2015, Conference Track
  Proceedings}.

\bibitem[{Sanh et~al.(2019)Sanh, Debut, Chaumond, and
  Wolf}]{DBLP:journals/corr/abs-1910-01108}
Victor Sanh, Lysandre Debut, Julien Chaumond, and Thomas Wolf. 2019.
\newblock Distilbert, a distilled version of {BERT:} smaller, faster, cheaper
  and lighter.
\newblock \emph{CoRR}.

\bibitem[{Sennrich et~al.(2016)Sennrich, Haddow, and
  Birch}]{DBLP:conf/acl/SennrichHB16a}
Rico Sennrich, Barry Haddow, and Alexandra Birch. 2016.
\newblock Neural machine translation of rare words with subword units.
\newblock In \emph{Proceedings of the 54th Annual Meeting of the Association
  for Computational Linguistics, {ACL} 2016, August 7-12, 2016, Berlin,
  Germany, Volume 1: Long Papers}. The Association for Computer Linguistics.

\bibitem[{Tucker(1966)}]{tucker1966some}
Ledyard~R Tucker. 1966.
\newblock Some mathematical notes on three-mode factor analysis.
\newblock \emph{Psychometrika}, 31(3):279--311.

\bibitem[{Vaswani et~al.(2017)Vaswani, Shazeer, Parmar, Uszkoreit, Jones,
  Gomez, Kaiser, and Polosukhin}]{DBLP:conf/nips/VaswaniSPUJGKP17}
Ashish Vaswani, Noam Shazeer, Niki Parmar, Jakob Uszkoreit, Llion Jones,
  Aidan~N. Gomez, Lukasz Kaiser, and Illia Polosukhin. 2017.
\newblock Attention is all you need.
\newblock In \emph{Advances in Neural Information Processing Systems 30: Annual
  Conference on Neural Information Processing Systems 2017, 4-9 December 2017,
  Long Beach, CA, {USA}}, pages 5998--6008.

\bibitem[{Wang et~al.(2020)Wang, Wu, Liu, Cai, Zhu, Gan, and Han}]{wang2020hat}
Hanrui Wang, Zhanghao Wu, Zhijian Liu, Han Cai, Ligeng Zhu, Chuang Gan, and
  Song Han. 2020.
\newblock Hat: Hardware-aware transformers for efficient natural language
  processing.

\bibitem[{Wang et~al.(2019)Wang, Li, Xiao, Zhu, Li, Wong, and
  Chao}]{DBLP:conf/acl/WangLXZLWC19}
Qiang Wang, Bei Li, Tong Xiao, Jingbo Zhu, Changliang Li, Derek~F. Wong, and
  Lidia~S. Chao. 2019.
\newblock Learning deep transformer models for machine translation.
\newblock In \emph{Proceedings of the 57th Conference of the Association for
  Computational Linguistics, {ACL} 2019, Florence, Italy, July 28- August 2,
  2019, Volume 1: Long Papers}, pages 1810--1822. Association for Computational
  Linguistics.

\bibitem[{Xiao et~al.(2012)Xiao, Zhu, Zhang, and Li}]{DBLP:conf/acl/XiaoZZL12}
Tong Xiao, Jingbo Zhu, Hao Zhang, and Qiang Li. 2012.
\newblock Niutrans: An open source toolkit for phrase-based and syntax-based
  machine translation.
\newblock In \emph{The 50th Annual Meeting of the Association for Computational
  Linguistics, Proceedings of the System Demonstrations, July 10, 2012, Jeju
  Island, Korea}, pages 19--24. The Association for Computer Linguistics.

\bibitem[{Yang et~al.(2019)Yang, Dai, Yang, Carbonell, Salakhutdinov, and
  Le}]{DBLP:conf/nips/YangDYCSL19}
Zhilin Yang, Zihang Dai, Yiming Yang, Jaime~G. Carbonell, Ruslan Salakhutdinov,
  and Quoc~V. Le. 2019.
\newblock Xlnet: Generalized autoregressive pretraining for language
  understanding.
\newblock In \emph{Advances in Neural Information Processing Systems 32: Annual
  Conference on Neural Information Processing Systems 2019, NeurIPS 2019, 8-14
  December 2019, Vancouver, BC, Canada}, pages 5754--5764.

\bibitem[{Zhang et~al.(2020)Zhang, Wang, Cao, Wei, Shan, Zhou, Reheman, Zhou,
  Zeng, Wang, Mu, Zhang, Liu, Zhou, Li, Li, Xiao, and
  Zhu}]{DBLP:conf/wmt/ZhangWCWSZRZZWM20}
Yuhao Zhang, Ziyang Wang, Runzhe Cao, Binghao Wei, Weiqiao Shan, Shuhan Zhou,
  Abudurexiti Reheman, Tao Zhou, Xin Zeng, Laohu Wang, Yongyu Mu, Jingnan
  Zhang, Xiaoqian Liu, Xuanjun Zhou, Yinqiao Li, Bei Li, Tong Xiao, and Jingbo
  Zhu. 2020.
\newblock The niutrans machine translation systems for {WMT20}.
\newblock In \emph{Proceedings of the Fifth Conference on Machine Translation,
  WMT@EMNLP 2020, Online, November 19-20, 2020}, pages 338--345. Association
  for Computational Linguistics.

\end{thebibliography}


\appendix

\section{Appendices}
\label{sec:appendix}

\noindent\textbf{Hyper-parameters.} We tune the learning rate and warmup steps in Phase 2 of WD. We use the grid search to select the learning rate in $[1\times10^{-4},3\times10^{-4},5\times10^{-4},7\times10^{-4},9\times10^{-4}]$ and warmup steps in $[1000,2000,3000,4000,5000]$ that have the best average BLEU performance in all validation sets.

\noindent\textbf{Datasets.} Detailed data statistics as well as the URLs of three machine translation tasks we used, including WMT16 English-Roman (En-Ro), NIST12 Chinese-English (Zh-En), and WMT14 English-German (En-De), are shown in \tab{tab:data}.

For En-Ro, the training set consists of 0.6M bilingual sentence pairs. The validation set \emph{newsdev-2016} contains 1999 pairs and the test set \emph{newstest-2016} contains 1999 pairs. For Zh-En, the training set consists of 1.8M bilingual sentence pairs. The validation set \emph{mt06} contains 1,664 pairs and the test set \emph{mt08} contains 1,357 pairs. For En-De, the training set consists of 4.5M bilingual sentence pairs. The validation set \emph{newstest-2013} contains 3,000 pairs and the test set \emph{newstest-2014} contains 3,003 pairs.

\noindent\textbf{Runtime.} To compare the average runtime for each approach, \tab{tab:result_transformer} shows the actual number of updates and runtime. For the baseline models (i.e., Teacher, \tinystu{} and \smallstu{}) and KD, we record their runtime in the Phase 1 entry because they only need to be trained once.

One can observe that in \tab{tab:result_transformer}, Phase 2 of WD generally consumes similar or less time as well as the number of updates than other approaches. This is because the model is already close to the optimum before the fine-tuning (Phase 2). \tab{tab:result_transformer} also shows that the number of updates in Phase 1 of WD is much less than other approaches, yet its training time is much longer. This phenomenon is more obvious in Transformer-deep models. This is because one step in Phase 1 of WD is 2.11$\times$ slower than in Phase 2 of WD.

\begin{table}[!t]
  \centering
  \setlength{\tabcolsep}{1.0mm}{
  \begin{tabular}{c|r|r|r|r|r|r}
  \hline
  \multicolumn{1}{c|}{\multirow{2}*{Lang.}}&
  \multicolumn{2}{c|}{Train} &
  \multicolumn{2}{c|}{Test} &
  \multicolumn{2}{c}{Valid}\\
  \cline{2-7}
  & \multicolumn{1}{c|}{Sent.} & \multicolumn{1}{c|}{Word} & \multicolumn{1}{c|}{Sent.} & \multicolumn{1}{c|}{Word} & \multicolumn{1}{c|}{Sent.} & \multicolumn{1}{c}{Word} \\
  \cline{1-7}
  \href{https://github.com/nyu-dl/dl4mt-nonauto}{En-Ro} & 0.6M & 33M & 1999 & 112K & 1999 & 118K \\
  \cline{1-7}
  \href{https://catalog.ldc.upenn.edu/}{Zh-En} & 1.8M & 115M & 1357 & 247K & 1664 & 280K \\
  \cline{1-7}
  \href{http://statmt.org/wmt14/translation-task.html}{En-De} & 4.5M & 262M & 3003 & 164K & 3000 & 156K \\
  \hline
  \end{tabular}
  \caption{Date statistics.}
  \label{tab:data}
  }
\end{table}

\noindent\textbf{Decoder.} We also investigate how WD's performance (on the validation set) and speed change given different decoder depths and widths. We choose the speed of WD to compute the speedup of different decoder depths and widths. Although the actual speedup of KD will not be exactly the same as the one of WD due to their different decoding results, they are close.

As shown in \tab{tab:shape}, WD is robust to different sized decoders, with both BLEU and speed significantly outperform KD. WD consistently outperforms KD by about 1 BLEU point under various decoder depths and widths. Interestingly, we find that pruning the layers degrades the performance more than shrinking its width, but it provides a higher speedup. Taking the student network with depth 2 and width 512 as an example, if we shrink the depth from 2 to 1, there is a decrease of 1.21 BLEU points in WD but with 1.12$\times$ speedup. When we shrink the width from 512 to 256, it leads to a moderate decrease of 0.59 BLEU points yet with only 1.06$\times$ speedup. This might be because layers are computed sequentially and wider matrices enjoy the parallel computation acceleration provided by modern GPUs.

\begin{table*}[t!]
  \centering
  \begin{tabular}{c|l|c|c|c|c|r|c|r|c}
  \hline
  &
  \multicolumn{1}{c|}{\multirow{2}*{System}}&
  \multicolumn{1}{c|}{\multirow{2}*{Depth}} &
  \multicolumn{1}{c|}{\multirow{2}*{Width}} &
  \multicolumn{1}{c|}{\multirow{2}*{Test}} &
  \multicolumn{1}{c|}{\multirow{2}*{Valid}} &
  \multicolumn{2}{c|}{Phase 1} &
  \multicolumn{2}{c}{Phase 2} \\
  \cline{7-10}
  & & & & & & \multicolumn{1}{c|}{\#Update} & \multicolumn{1}{c|}{Time} & \multicolumn{1}{c|}{\#Update} & \multicolumn{1}{c}{Time} \\
  \hline
  \multirow{7}{*}{\rotatebox{90}{WMT16 En-Ro}} &
  \multirow{1}{*}{Teacher (base)} & 6 & 512 & 31.64 & 32.07 & 70K & 0.06 & - & - \\
  \cline{2-10}
  & \multirow{1}{*}{\tinystu{}} & 1 & 256 & 29.65 & 29.73 & 70K & 0.03 & - & - \\
  & \multirow{1}{*}{\quad + KD} & 1 & 256 & 30.03 &29.98 & 70K & 0.03 & - & - \\
  & \multirow{1}{*}{\quad + WD} & 1 & 256 & 30.89 &30.89 & 47K & 0.04 & 70K & 0.03 \\
  \cline{2-10}
  & \multirow{1}{*}{\smallstu{}} & 2 & 512 & 31.22 & 31.19 & 70K & 0.04 & - & - \\
  & \multirow{1}{*}{\quad + KD} & 2 & 512 & 30.97 & 30.77 & 70K & 0.04 & - & - \\
  & \multirow{1}{*}{\quad + WD} & 2 & 512 & 31.65 & 31.27 & 47K & 0.06 & 70K & 0.04 \\
  \hline
  \multirow{7}{*}{\rotatebox{90}{NIST12 Zh-En}} &
  \multirow{1}{*}{Teacher (base)} & 6 & 512 & 45.14 & 51.91 & 30K & 0.08 & - & - \\
  \cline{2-10}
  & \multirow{1}{*}{\tinystu{}} & 1 & 256 & 41.90 &48.28 & 30K & 0.05 & - & - \\
  & \multirow{1}{*}{\quad + KD} & 1 & 256 & 42.78 & 49.71 & 30K & 0.05 & - & - \\
  & \multirow{1}{*}{\quad + WD} & 1 & 256 & 44.60 &51.56 & 20K & 0.07 & 30K & 0.05 \\
  \cline{2-10}
  & \multirow{1}{*}{\smallstu{}} & 2 & 512 & 44.30 & 50.83 & 30K & 0.06 & - & - \\
  & \multirow{1}{*}{\quad + KD} & 2 & 512 & 44.89 & 51.87 & 30K & 0.06 & - & - \\
  & \multirow{1}{*}{\quad + WD} & 2 & 512 & 46.20 & 53.04 & 20K & 0.09 & 30K & 0.06 \\
  \hline
  \multirow{21}{*}{\rotatebox{90}{WMT14 En-De}} &
  \multirow{1}{*}{Teacher (base)} & 6 & 512 & 27.47 & 26.79 & 100K & 0.24 & - & - \\
  \cline{2-10}
  & \multirow{1}{*}{\tinystu{}} & 1 & 256 & 24.62 & 24.88 & 100K & 0.14 & - & - \\
  & \multirow{1}{*}{\quad + KD} & 1 & 256 & 26.51 & 26.01 & 100K & 0.14 & - & - \\
  & \multirow{1}{*}{\quad + WD} & 1 & 256 & 27.12 & 26.42 & 50K & 0.18 & 80K & 0.11 \\
  \cline{2-10}
  & \multirow{1}{*}{\smallstu{}} & 2 & 512 & 26.68 & 26.07 & 100K & 0.19 & - & - \\
  & \multirow{1}{*}{\quad + KD} & 2 & 512 & 27.47 & 26.54 & 100K & 0.19 & - & - \\
  & \multirow{1}{*}{\quad + WD} & 2 & 512 & 28.18 & 26.97 & 50K & 0.25 & 80K & 0.15 \\
  \cline{2-10}
  & \multirow{1}{*}{Teacher (big)} & 6 & 1024 & 29.11 & 27.66 & 200K & 1.71 & - & - \\
  \cline{2-10}
  & \multirow{1}{*}{\tinystu{}} & 1 & 512 & 25.83 & 25.33 & 200K & 0.58 & - & - \\
  & \multirow{1}{*}{\quad + KD} & 1 & 512 & 27.70 & 26.52 & 200K & 0.58 & - & - \\
  & \multirow{1}{*}{\quad + WD} & 1 & 512 & 28.60 & 26.83 & 67K & 0.57 & 100K & 0.29 \\
  \cline{2-10}
  & \multirow{1}{*}{\smallstu{}} & 2 & 1024 & 27.62 & 26.78 & 200K & 0.79 & - & - \\
  & \multirow{1}{*}{\quad + KD} & 2 & 1024 & 29.01 & 27.54 & 200K & 0.79 & - & - \\
  & \multirow{1}{*}{\quad + WD} & 2 & 1024 & 29.52 & 27.97 & 67K & 0.55 & 100K & 0.40 \\
  \cline{2-10}
  & \multirow{1}{*}{Teacher (deep)} & 6 & 512 & 29.43 & 27.82 & 60K & 0.67 & - & - \\
  \cline{2-10}
  & \multirow{1}{*}{\tinystu{}} & 1 & 256 & 26.34 & 26.05 & 60K & 0.57 & - & - \\
  & \multirow{1}{*}{\quad + KD} & 1 & 256 & 29.36 & 27.39 & 60K & 0.57 & - & - \\
  & \multirow{1}{*}{\quad + WD} & 1 & 256 & 29.92 & 27.99 & 30K & 1.51 & 30K & 0.29 \\
  \cline{2-10}
  & \multirow{1}{*}{\smallstu{}} & 2 & 512 & 28.06 & 26.51 & 60K & 0.60 & - & - \\
  & \multirow{1}{*}{\quad + KD} & 2 & 512 & 29.83 & 28.02 & 60K & 0.60 & - & - \\
  & \multirow{1}{*}{\quad + WD} & 2 & 512 & 30.77 & 28.33 & 30K & 1.53 & 30K & 0.30 \\
  \hline
  \end{tabular}
  \caption{Results of Transformer on different tasks (Time is measured by GPU days).}
  \label{tab:result_transformer}
\end{table*}

\begin{table*}[t!]
  \centering
    \setlength{\tabcolsep}{2.5mm}{
    \begin{tabular}{c|c|c|c|c|c|c|c|c}
      \hline
      \multicolumn{1}{c|}{\multirow{2}{*}{\diagbox{D}{W}}} &
      \multicolumn{4}{c|}{256} &
      \multicolumn{4}{c}{512} \\
      \cline{2-9}
      &
      \multicolumn{1}{c|}{KD} &
      \multicolumn{1}{c|}{WD} &
      \multicolumn{1}{c|}{$\mathrm{\Delta}_{\text{BLEU}}$} &
      \multicolumn{1}{c|}{Speedup} &
      \multicolumn{1}{c|}{KD} &
      \multicolumn{1}{c|}{WD} &
      \multicolumn{1}{c|}{$\mathrm{\Delta}_{\text{BLEU}}$} &
      \multicolumn{1}{c}{Speedup} \\
      \hline
      1 & 49.71 & 51.56 & +1.85 & 2.80$\times$ & 50.89 & 51.83 & +0.94 & 2.53$\times$ \\
      2 & 51.25 & 52.45 & +1.20 & 2.12$\times$ & 51.87 & 53.04 & +1.17 & 2.25$\times$ \\
      3 & 51.52 & 52.49 & +0.97 & 1.78$\times$ & 52.46 & 52.81 & +0.35 & 1.68$\times$ \\
      4 & 51.41 & 52.42 & +1.01 & 1.62$\times$ & 52.07 & 53.66 & +1.59 & 1.56$\times$ \\
      5 & 51.27 & 52.71 & +1.44 & 1.33$\times$ & 52.07 & 52.74 & +0.67 & 1.30$\times$ \\
      6 & 50.79 & 52.65 & +1.86 & 1.18$\times$ & 51.91 & 53.09 & +1.18 & 1.02$\times$ \\
      \hline
    \end{tabular}
    \caption{BLEU and speed vs. decoder depth and width (Transformer-base, NIST12 Zh-En).}
    \label{tab:shape}
    }
\end{table*}

\begin{figure}[!t]
  \centering
  \begin{tikzpicture}
    \begin{axis}[
      width=\linewidth,height=0.56\linewidth,
      yticklabel style={/pgf/number format/fixed,/pgf/number format/precision=1},
      ylabel={Loss},
      ylabel near ticks,
      xlabel={\#Update ($\times 10^3$)},
      xlabel near ticks,
      enlargelimits=0.05,
      xmajorgrids=true,
      ymajorgrids=true,
      grid style=dashed,
      xtick=data,
      every tick label/.append style={font=\small},
      label style={font=\small},
      ylabel style={yshift=5pt},
      legend style={font=\small,inner sep=3pt},
      legend image post style={scale=1},
      legend columns=1,
      legend cell align={left},
    ]
      \addplot [lyygreen,thick,mark=*] coordinates {
        (2,5.68) (5,4.57) (10,4.32) (15,3.88) (20,3.76) (25,3.70) (30,3.73)
      };
      \addlegendentry{KD}
      \addplot [lyyred,thick,mark=square*] coordinates {
        (2,4.27) (5,4.03) (10,3.74) (15,3.62) (20,3.55) (25,3.54) (30,3.49)
      };
      \addlegendentry{WD (Phase 1)}
      \addplot [lyyblue,thick,mark=triangle*] coordinates {
        (2,3.62) (5,3.54) (10,3.64) (15,3.55) (20,3.45) (25,3.36) (30,3.34)
      };
      \addlegendentry{WD (Phase 2)}
      \addplot [lyygreen,dashed,mark=*] coordinates {
        (4,5.29) (8,4.59) (12,4.41) (16,4.32) (20,4.26) (24,4.25) (28,4.21) (30,4.21)
      };
      \addplot [lyyred,dashed,mark=square*] coordinates {
        (4,4.39) (8,4.26) (12,4.21) (16,4.20) (20,4.18) (24,4.17) (28,4.16) (30,4.16)
      };
      \addplot [lyyblue,dashed,mark=triangle*] coordinates {
        (4,4.22) (8,4.17) (12,4.15) (16,4.13) (20,4.14) (24,4.14) (28,4.13) (30,4.13)
      };
    \end{axis}
  \end{tikzpicture}
  \caption{Train (solid)/valid (dash) loss of \smallstu{}.}
  \label{fig:curve}
\end{figure}
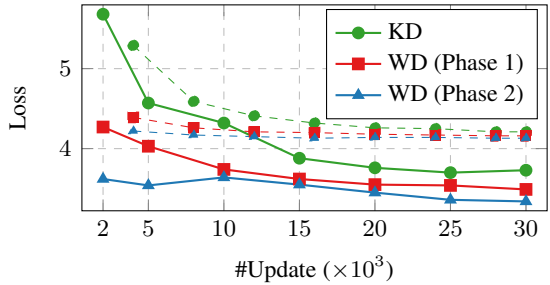

\noindent\textbf{Loss.} In \tab{tab:result_transformer}, we observe that WD generates student networks that are superior to KD. We believe that this is because WD converges to a better optimum. To examine this hypothesis, we study its loss in \fig{fig:curve}. As can be seen, WD does obtain much lower train and valid losses than KD. We also see that Phase 1 already outperforms KD at the end. Given the fact that Phase 1 does the initialization job for Phase 2 and Phase 2 is KD exactly, the way WD works can be treated as providing a good start.

\end{document}